%% file: main.tex
\documentclass[10pt,twocolumn,letterpaper]{article}

\usepackage[pagenumbers]{cvpr}   

\input{preamble}

\definecolor{cvprblue}{rgb}{0.21,0.49,0.74}
\begin{document}

\title{AI-Generated Image Quality Assessment Based on Task-Specific Prompt and Multi-Granularity Similarity}

\author{
		Jili Xia\IEEEauthorrefmark{1},
		Lihuo He\IEEEauthorrefmark{1*},
		Fei Gao\IEEEauthorrefmark{2}, 
		Kaifan Zhang\IEEEauthorrefmark{1}, 
		Leida Li\IEEEauthorrefmark{3},
		and
            \vspace{4mm}
		Xinbo Gao\IEEEauthorrefmark{1,4} \\
	\IEEEauthorrefmark{1}School of Electronic Engineering, Xidian University, Xi’an, China \\
	\IEEEauthorrefmark{2}Hangzhou Institute of Technology, Xidian University, Xi’an, China \\
        \IEEEauthorrefmark{3}School of Artificial Intelligence, Xidian University, Xi’an, China \\
	\IEEEauthorrefmark{4}Chongqing University of Posts and Telecommunications, Chongqing, China
}

\maketitle
\input{sec/0_abstract}    
\input{sec/1_intro}

\input{sec/2_related}
\input{sec/3_method}
\input{sec/4_experiments}
\input{sec/5_conclusion}
{
    \small
    \bibliographystyle{ieeenat_fullname}
    \bibliography{main}
}


\end{document}

%% file: preamble.tex
\newcommand{\IEEEauthorrefmark}[1]{\textsuperscript{#1}}
\usepackage{multirow, makecell}
\usepackage{algorithm} 
\usepackage{algorithmicx} 
\usepackage[dvipsnames]{xcolor}
\usepackage[pagebackref,breaklinks,colorlinks,allcolors=cvprblue]{hyperref}


%% file: sec/0_abstract.tex
\begin{abstract}
	
Recently, AI-generated images (AIGIs) created by given prompts (initial prompts) have garnered widespread attention. Nevertheless, due to technical nonproficiency, they often suffer from poor perception quality and Text-to-Image misalignment. Therefore, assessing the perception quality and alignment quality of AIGIs is crucial to improving the generative model's performance. Existing assessment methods overly rely on the initial prompts in the task prompt design and use the same prompts to guide both perceptual and alignment quality evaluation, overlooking the distinctions between the two tasks. To address this limitation, we propose a novel quality assessment method for AIGIs named TSP-MGS, which designs task-specific prompts and measures multi-granularity similarity between AIGIs and the prompts. Specifically, task-specific prompts are first constructed to describe perception and alignment quality degrees separately, and the initial prompt is introduced for detailed quality perception. Then, the coarse-grained similarity between AIGIs and task-specific prompts is calculated, which facilitates holistic quality awareness. In addition, to improve the understanding of AIGI details, the fine-grained similarity between the image and the initial prompt is measured. Finally, precise quality prediction is acquired by integrating the multi-granularity similarities. Experiments on the commonly used AGIQA-1K and AGIQA-3K benchmarks demonstrate the superiority of the proposed TSP-MGS. 

\end{abstract}

%% file: sec/1_intro.tex
\section{Introduction}
\label{sec:intro}

\begin{figure}[htbp]
	\centering
    \subfloat[]{
	\begin{minipage}[t]{.32\linewidth}
		\centering
		\includegraphics[width=.96\linewidth]{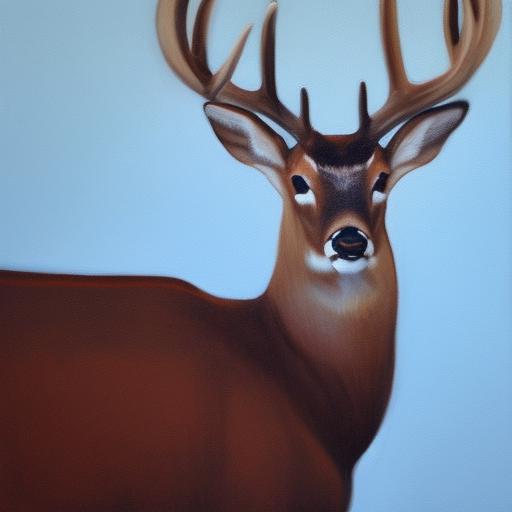}
		\captionsetup{justification=raggedright, singlelinecheck=false, font=footnotesize} 
		\setlength{\abovecaptionskip}{2pt}
		\caption*{a portrait painting of \textcolor{RawSienna}{a buck in suit}}
	\end{minipage}
	\begin{minipage}[t]{.32\linewidth}
		\includegraphics[width=.96\linewidth]{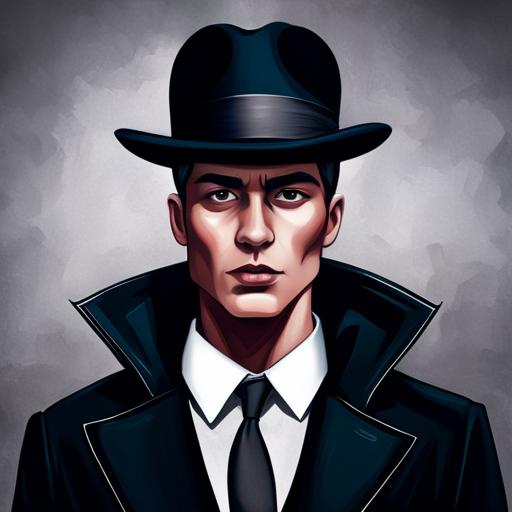}
		\captionsetup{justification=raggedright, singlelinecheck=false, font=footnotesize} 
		\setlength{\abovecaptionskip}{2pt}
		\caption*{black \textcolor{RawSienna}{mickey mouse skull}}
	\end{minipage}
	\begin{minipage}[t]{.32\linewidth}
		\includegraphics[width=.96\linewidth]{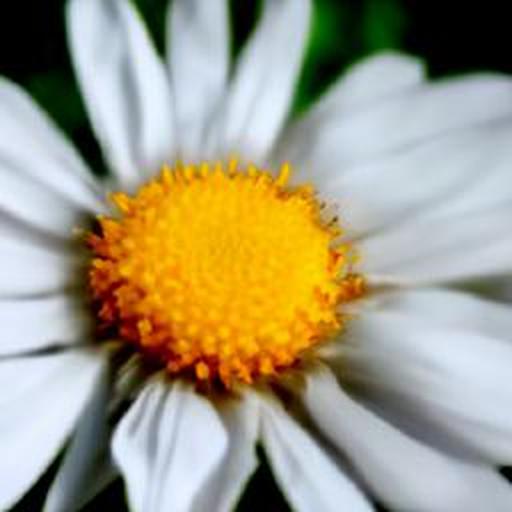}
		\captionsetup{justification=raggedright, singlelinecheck=false, font=footnotesize} 
		\setlength{\abovecaptionskip}{2pt}
		\caption*{photography of \textcolor{RawSienna}{a happy blonde girl}}
	\end{minipage}
    \label{fig:1a}
    }

    \subfloat[]{
	\begin{minipage}[t]{.32\linewidth}
		\centering
		\includegraphics[width=.96\linewidth]{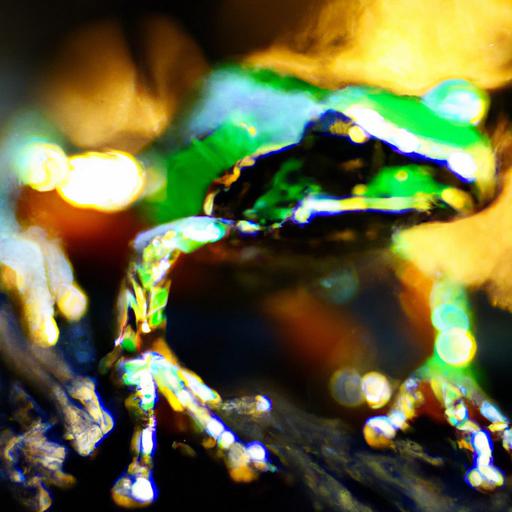}
		\captionsetup{justification=raggedright, singlelinecheck=false, font=footnotesize} 
		\setlength{\abovecaptionskip}{2pt}	
		\caption*{a cybertronic metallic green tree frog}
	\end{minipage}
	\begin{minipage}[t]{.32\linewidth}
		\includegraphics[width=.96\linewidth]{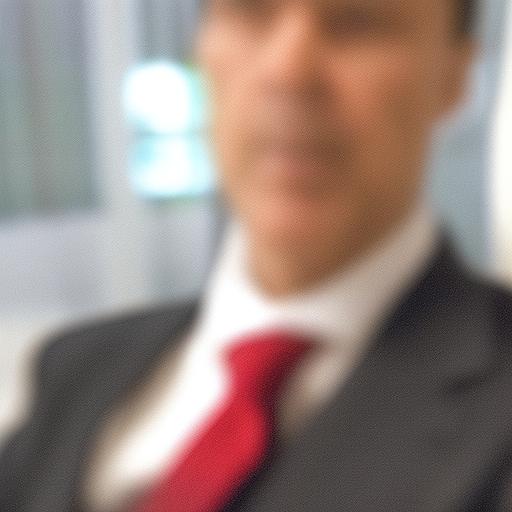}
		\captionsetup{justification=raggedright, singlelinecheck=false, font=footnotesize} 
		\setlength{\abovecaptionskip}{2pt}
		\caption*{man in his late 40s wearing a suit}
	\end{minipage}
	\begin{minipage}[t]{.32\linewidth}
		\includegraphics[width=.96\linewidth]{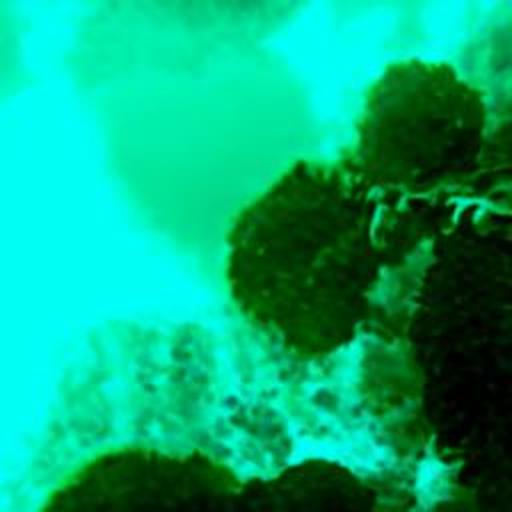}
		\captionsetup{justification=raggedright, singlelinecheck=false, font=footnotesize} 
		\setlength{\abovecaptionskip}{2pt}
		\caption*{an underwater photo of coral in the ocean}
	\end{minipage}
    \label{fig:1b}
    }
    
     \subfloat[]{
	\begin{minipage}[t]{.32\linewidth}
		\centering
		\includegraphics[width=.96\linewidth]{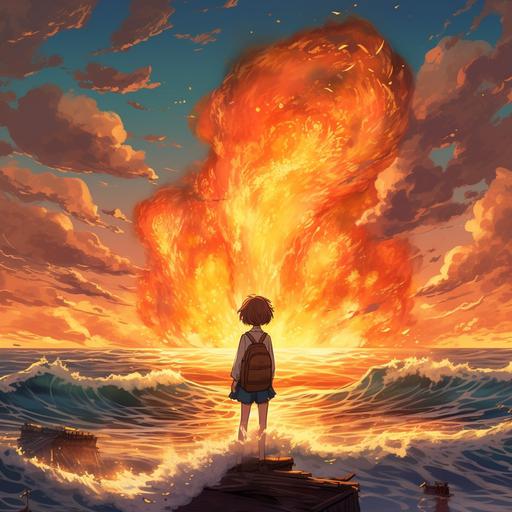}
		\captionsetup{justification=raggedright, singlelinecheck=false, font=footnotesize} 
		\setlength{\abovecaptionskip}{2pt}
		\caption*{ocean of canvas that catches fire}
	\end{minipage}
	\begin{minipage}[t]{.32\linewidth}
		\includegraphics[width=.96\linewidth]{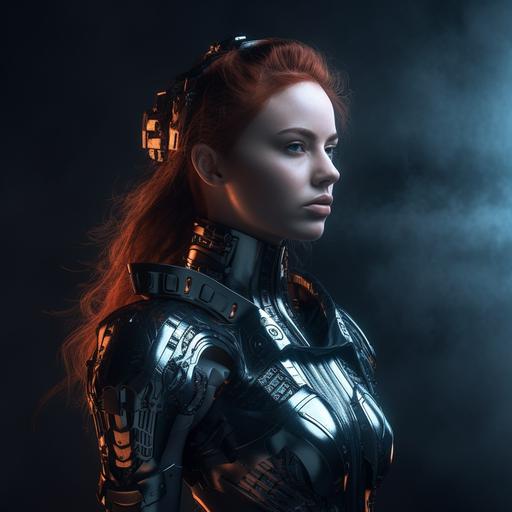}
		\captionsetup{justification=raggedright, singlelinecheck=false, font=footnotesize} 
		\setlength{\abovecaptionskip}{2pt}
		\caption*{portrait of beautiful armored girl}
	\end{minipage}
	\begin{minipage}[t]{.32\linewidth}
		\includegraphics[width=.96\linewidth]{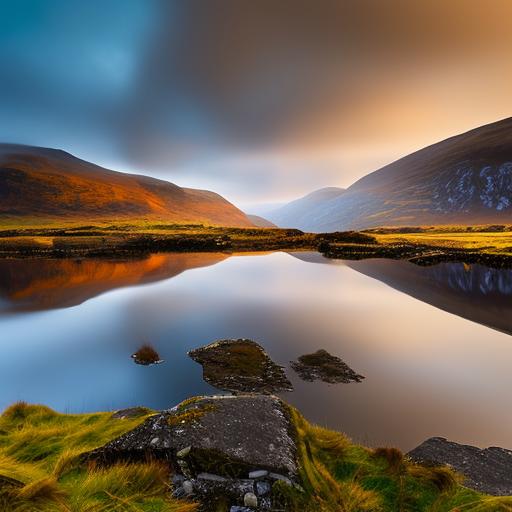}
		\captionsetup{justification=raggedright, singlelinecheck=false, font=footnotesize} 
		\setlength{\abovecaptionskip}{2pt}
		\caption*{scottish highlands at dawn}
	\end{minipage}
    }
	 \caption{Illustration of quality degradations of AIGIs. (a) AIGIs do not align with the initial prompts. (b) AIGIs suffer from poor perception quality. (c) AIGIs present high alignment quality and perception quality.}\label{fig:1}
\end{figure}

Artificial Intelligence-Generated Content (AIGC), especially AI-generated images (AIGIs), attracts wide attention across various fields due to its flexibility and convenience. Recent advancements in Text-to-Image (T2I) generation models, such as Generative Adversarial Networks (GANs)~\cite{Stackgan, Attngan}, regression-based models~\cite{Cogview, regression02}, and diffusion-based models~\cite{Diffusion01, Diffusion02}, allow diverse AI-generated images to be available. However, the quality of AI-generated images is inconsistent due to the instability of generation techniques, limiting the broader applications of these images. Therefore, developing effective quality assessment models is essential for improving the generation model's capabilities and selecting high-quality AI-generated images.

Over recent decades, general-purpose image quality assessment (IQA) methods have made substantial progress, which primarily focus on images degraded by artificial distortions~\cite{live,csiq,kadid-10k}, such as compression, noise, and blurring, as well as images captured in the wild~\cite{livec,koniq-10k,spaq,LIVEFB}. They evaluate perception quality based on the distortion properties extracted from global images or local image patches~\cite{BRISQUE,deepiqa,tres,re-iqa,LoDa}. However, degradations of AIGIs are unique~\cite{aigc1k,aigc3k}, making these methods inapplicable. Specifically, AIGIs usually do not align with the initial prompts due to the poor prompt comprehension of generative models, as shown in \cref{fig:1a}. Additionally, AIGIs often exhibit low perception quality for the limited generation capabilities of generative models, as illustrated in \cref{fig:1b}. Therefore, an effective method is required to evaluate both the alignment and perception quality of AIGIs.        

To this end, advanced AIGC image quality assessment (AIGCIQA) methods have been proposed over the last few years and can be broadly categorized into three types. The first type~\cite{ImageReward, HPS, PickScore,aigc3k} evaluates human preference for AIGIs but falls short in proving a thorough understanding of quality. The second type~\cite{IPCE,PCQA,CLIP-AGIQA} employs vision-language models (VLMs), such as CLIP~\cite{CLIP}, to evaluate the alignment and perception quality of AIGIs. They construct task prompts based on the initial prompts and use the same task prompts to guide the above evaluation learning. However, such task prompts tend to favor alignment quality prediction, which reduces the model’s effectiveness in predicting perception quality. The third type~\cite{IP-IQA,MoE-AGIQA} introduces an additional single-modal image encoder to boost perception quality evaluation while using VLMs for alignment quality assessment. Although this design is effective, it increases the model’s complexity. 

To address the above challenges, we propose an advanced AIGCIQA method named TSP-MGS. To resolve task ambiguity caused by shared task prompts, we design task-specific prompts for alignment and perception quality predictions to describe the quality degree, which enhances the model's perception of the two tasks. Considering that initial prompts provide rich descriptions of AIGI details, we introduce them to guide the model's detail awareness. The above methods only consider the coarse-grained similarity between AIGIs and prompts, neglecting the important detail distortions. To address this, we present the multi-granularity similarity measurement for a comprehensive quality evaluation. Specifically, the sentence-level (coarse-grained) similarity between AIGIs and their task-specific prompts is measured to capture overall quality-aware representations. Then, the word-level (fine-grained) similarity between AIGIs and their initial prompts is calculated to learn detailed quality-aware representations. By integrating the coarse-grained and fine-grained similarities, we achieve precise quality prediction of AIGIs.
                   
The main contributions of this paper are summarized as follows:
\begin{itemize}  
	\item We design task-specific prompts to describe alignment and perception quality degree, improving the model's awareness of each task.      
	\item We compute multi-granularity similarities between AIGIs and their prompts, promoting holistic and detailed awareness of quality representation.  
    \item Integrating the task-specific prompts and the multi-granularity similarity, we propose an effective AIGCIQA method named TSP-MGS, which achieves state-of-the-art quality predictions on AGIQA-1K and AGIQA-3K benchmarks.    
\end{itemize}


%% file: sec/2_related.tex
\section{Related Work}
\label{sec:related}

\begin{figure*}[!t]
    \centering
    \begin{minipage}[t]{1.\linewidth}
        \centering
        \includegraphics[width=1\linewidth]{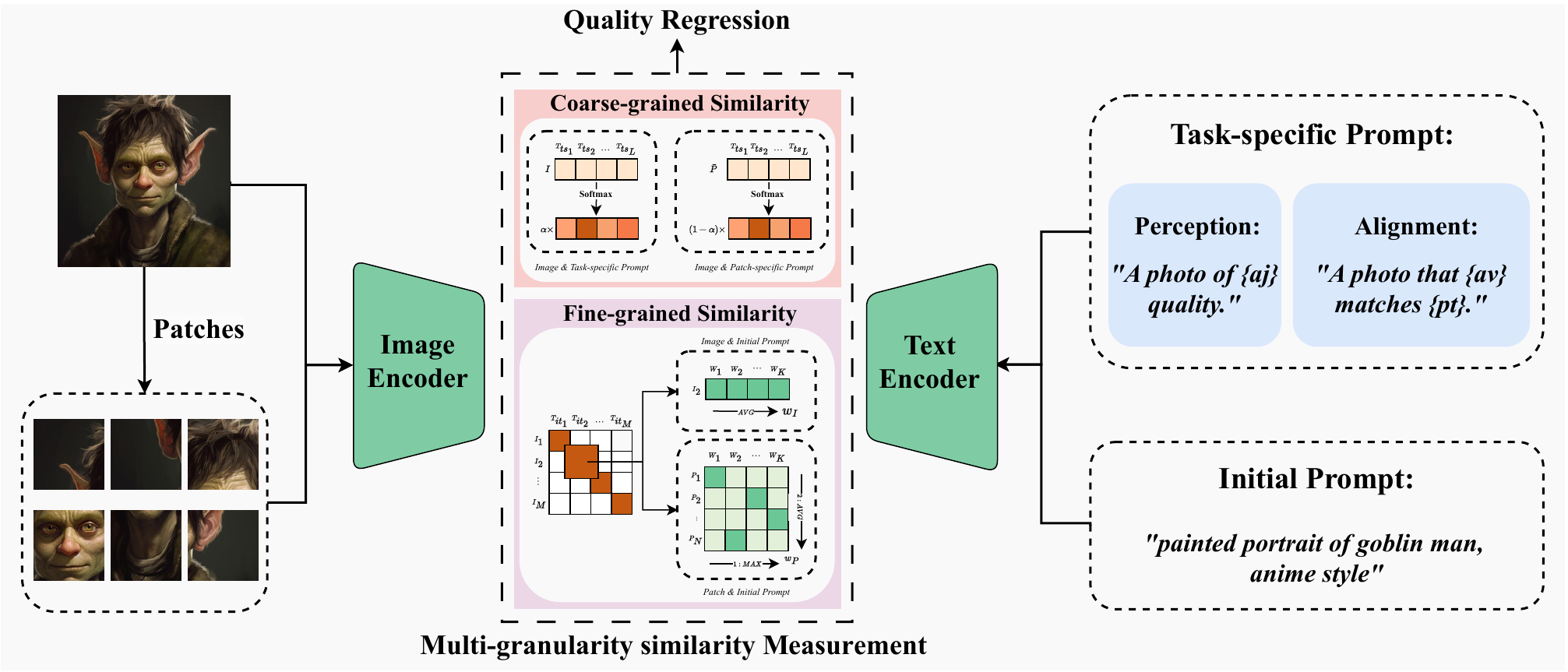}
    \end{minipage}
    \caption{The pipeline of the proposed method. First, we construct task-specific prompts to describe the degree of perception and alignment quality, while introducing the initial prompt for image content understanding. Using the CLIP model, we then extract image features from both the full image and cropped patches, along with text features from the task-specific prompts and the initial prompt. For a comprehensive quality perception and content understanding, we calculate multi-granularity similarities, \textit{i.e.,} coarse-grained similarity and fine-grained similarity, between the image features and text features. Finally, we integrate these similarities for a precise quality prediction.}\label{fig:pipeline}
\end{figure*}

This section reviews representative general-purpose IQA methods and existing artificial intelligence-generated content IQA methods.

\subsection{General-Purpose Image Quality Assessment}

General-purpose IQA methods typically evaluate synthetically or authentically distorted images~\cite{live,csiq,kadid-10k,livec,koniq-10k,spaq,LIVEFB}, with an important part of reliable degradation-aware feature extraction. 

Traditional IQA methods~\cite{BRISQUE,BLIINDS-II,MKL,ILNIQE,LOCRUE} often leverage hand-crafted features extracted from the spatial or transform domains to characterize distortions. However, these features are limited in capturing complex distortions, such as mixed distortions and authentic distortions, which significantly impair the performance of traditional methods. Convolutional neural network (CNN)-based IQA methods~\cite{deepiqa,hyperiqa,dbcnn,dacnn,oln,fuzzy} improve the ability to assess complex distortions by leveraging the automatic feature extraction capabilities of CNN models. Moreover, they demonstrate superior adaptability to diverse distortions and have better generalization capacity. However, they fall short in combining contextual information, which is crucial for accurate quality assessment. In contrast, Vision Transformer (ViT)-based IQA methods~\cite{TranSLA,tres,MANIQA,wang,TempQT} address this limitation by using self-attention mechanisms that enable the model to build long-range dependency among image patches.

VLMs aim to model image-text correlations to facilitate zero-shot visual recognition~\cite{survey}. Owing to the utilization of contrastive learning~\cite{CLIP} and pre-training on diverse image-text datasets~\cite{CLIP,FILIP,GLIP}, VLMs exhibit excellent generalization capability and transferability, which has made them widely applicable in vision tasks, such as image classification~\cite{CLIP, FILIP}, segmentation~\cite{GroupViT,Segclip}, and detection~\cite{RegionCLIP,GLIP}. Inspired by this, the exploration of applying VLMs to IQA is attracting increased attention. Current researches focus on designing suitable textual descriptions~\cite{CLIPiqa, LIQE} and constructing datasets with quality description~\cite{qbench,qinstruct}. Multimodal IQA methods reduce the model's reliance on extensive manual annotations and provide more intuitive explanations of image quality, enhancing users' understanding of image distortions.                      

\subsection{AIGC Image Quality Assessment}
AIGIs~\cite{aigc1k,aigc3k} are generated based on their initial prompts, however, limitations in the text comprehension or image generation of generative models often result in T2I misalignment or poor perception quality. Consequently, existing AIGCIQA methods are primarily designed to address these two challenges. 

Some methods~\cite{ImageReward, HPS, PickScore} focus on learning human preference for AIGIs. However, they lack thorough quality perception. Other methods focus on predicting quality scores. Peng \textit{et al.}~\cite{IPCE} evaluated the alignment quality by employing the CLIP to measure the similarity between AIGIs and their prompts. Then, they transformed the measured similarity into precise quality scores. Fang \textit{et al.}~\cite{PCQA} mixed image features and prompt features derived from hybrid text encoders, enhance the model's text comprehension capabilities and quality assessment performance. These methods manually design prompts, leading to limited flexibility. To handle this limitation, Fu \textit{et al.}~\cite{CLIP-AGIQA} introduced learnable visual and textual prompts. Nevertheless, these methods emphasize alignment quality prediction but ignore perception quality. 

To this end, Yang \textit{et al.}~\cite{MoE-AGIQA} adopted an image encoder to extract perception degradation features and a VLMs to learn semantic-aware features. Then, they fused these features using a designed cross-attention module to achieve a comprehensive quality assessment. Unlike this, Yu \textit{et al.}~\cite{SF-IQA} extracted and fused multi-layer perception degradation features of the image, and they integrated the perception scores and alignment scores during the regression stage. Although these methods consider perception and alignment features simultaneously, an additional image encoder is needed for perception feature extraction, increasing the model's complexity while ignoring the extra information provided by text prompts.                      

%% file: sec/3_method.tex
\section{Method}
\label{sec:method}

In this section, we first illustrate the overall framework of the proposed method and briefly review CLIP~\cite{CLIP}. Then, we detail text prompt construction, multi-granularity similarity measurement, and quality regression.  

\subsection{Overall Framework}
The overall framework of our method is illustrated in \cref{fig:pipeline}. It adopts the CLIP~\cite{CLIP} as the baseline, which consists of an image encoder $f_{img}\left ( \cdot \right )$, a text encoder $f_{txt}\left ( \cdot \right )$, and the cosine similarity measurement. Given an image $I$ and the corresponding prompt $T$, the image feature $F_{I}$ and prompt feature $F_{T}$ can be formulated as
\begin{equation}
	F_{I}=f_{img}\left ( I \right ), F_{T}=f_{txt}\left ( T \right )
 	\label{eq:1}
\end{equation} 

Then, cosine similarity $S(F_{I}, F_{T})$ is calculated to measure the matching degree between $I$ and $T$, which can be expressed as
\begin{equation}
	S(F_{I}, F_{T})=\frac{F_{I}\odot F_{T}}{\left\|F_{I} \right\|\cdot \left\|F_{T} \right\|},
	\label{eq:2}
\end{equation} 
where $\odot$ is the vector dot-product and $\left\|\cdot \right\|$ means the $\ell_2$ norm. 

\begin{table*}[!t]
    \centering
    \setlength{\tabcolsep}{8mm}
    \begin{tabular}{cc}
        \toprule
        Explanation &Notation\\
        \midrule
        Image feature & $F_{I_r}$ \\
        Patch feature & $F_{P_i}$, $i \in \left\{1, 2,\cdots, N\right\}$ indexes the patch \\
        Word feature & $F_{W_k}$, $k \in \left\{1, 2,\cdots, K\right\}$ indexes the word of the initial prompt \\
        Task-specific prompt feature & $F_{{T_{ts}}_j}$, $j \in \left\{1, 2,\cdots, L\right\}$ indexes the quality level of the image \\
        \bottomrule
    \end{tabular}
    \caption{Explanation of Notations.}
    \label{tab:notation}
\end{table*} 

\begin{figure}[t]
	\centering
	\begin{minipage}[t]{1.\linewidth}
		\centering
		\includegraphics[trim=.6cm 0 .2cm  0, clip, width=.95\linewidth]{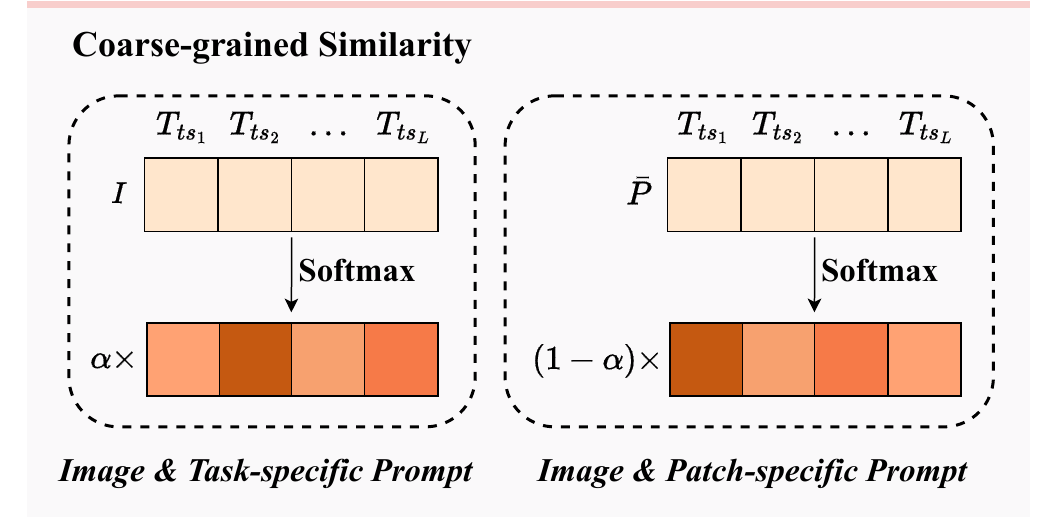}
	\end{minipage}
	\caption{Coarse-grained similarity measurement.  }\label{fig:3}
\end{figure}

\begin{figure}[tbp]
	\centering
	\begin{minipage}[t]{1.\linewidth}
		\centering
		\includegraphics[width=.95\linewidth]{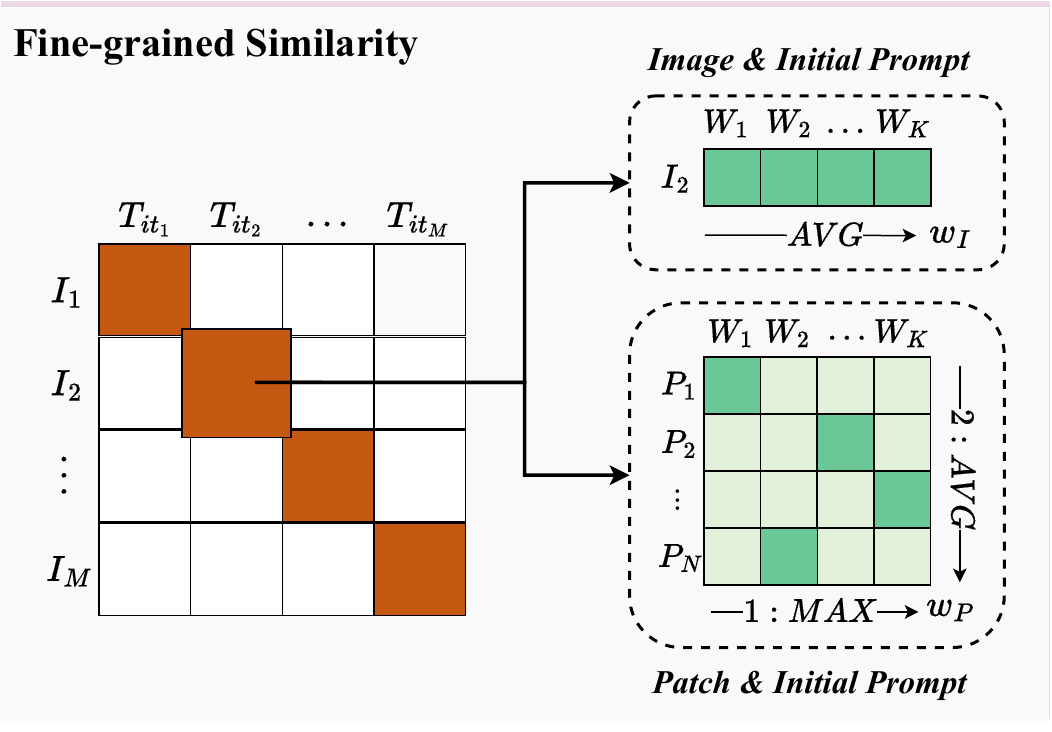}
	\end{minipage}
	\caption{Fine-grained similarity measurement.}\label{fig:4}
\end{figure}

The proposed method consists of three main steps: (1) feature extraction. It uses the resized AIGI $I_r$ and $N$ patches $\textit{\textbf{P}}=\left\{ P_i|i\in \left\{ 1,2,\cdots, N\right\}\right\}$ as the inputs of the image encoder, with the output features denoted as $F_{I_r}$ and $F_{P_i}$, respectively. Besides, the task-specific prompt $T_{ts}$ and initial prompt $T_{it}$ are used as the input of the text encoder, with their features represented as $F_{T_{ts}}$ and $F_{T_{it}}$, respectively; (2) multi-granularity similarity measurement. The coarse-grained similarity between the task-specific prompt and the image, as well as between the task-specific prompt and the patches is calculated. This enables a holistic quality perception. Moreover, the fine-grained similarity between the initial prompt and the image, as well as the initial prompt and the patches is measured. This provides a detailed quality understanding; (3) quality regression. It predicts an accurate AIGI quality by integrating the multi-granularity similarities.

\subsection{Text Prompt}
\label{sec:3.2}
As shown in \cref{fig:1}, alignment quality and perception quality of an AIGI do not strictly exhibit a positive correlation, demonstrating that using the same prompt for both quality evaluations may lead to ambiguous semantic guidance. To this end, we construct task-specific prompts to enhance the model's task-aware ability for alignment and perception quality evaluation.             
\paragraph{Task-Specific Prompt}
For alignment quality evaluation, we adopt an alignment-specific prompt to describe the alignment degree between AIGI and its initial prompt $\left\{ \textit{pt} \right\}$~\cite{IPCE}, denoted as
\begin{align}
	\text{}{T^{adv}_{ts}:\ ``A\ photo\ that\ \left\{ \textit{adv} \right\}\ matches\ \left\{ \textit{pt} \right\}."} 
	\nonumber
\end{align}
where $\textit{adv}$ is in $\left [badly, \ poorly,\ fairly, \ well, \ perfectly \right ]$.

For perception quality evaluation, we adopt a perception-specific prompt to describe the degree of visual quality. Inspired by ~\cite{CLIPiqa, LIQE}, we employ and compare two different text descriptions, which are shown as follows
\begin{align}
	&\text{}{T^{ant}_{ts}:\ ``\left\{ \textit{ant} \right\}\ photo."}
	\nonumber \\
	&\text{}{T^{adj}_{ts}:\ ``A\ photo\ of \ \left\{ \textit{adj} \right\} \ quality."}
	\nonumber 
\end{align}   
where $\textit{ant}$ is antonym description, which is one of $\left[bad, \ good \right ]$~\cite{CLIPiqa}, $\textit{adj}$ means adjective description selected from $\left [bad, \ poor,\ fair, \ good, \ perfect \right ]$~\cite{LIQE}.

Task-specific prompts, \textit{i.e.,} alignment-specific and perception-specific prompts, guide the model to form an overall quality identification of AIGIs. To further be aware of detailed image quality, we also include initial prompts as the input of the text encoder, denoted as $\text{}{T_{it}}$. 

\subsection{Multi-granularity Similarity}
We measure the coarse-grained similarity, shown in~\cref{fig:3}, between AIGIs and task-specific prompts (sentence level) to improve the model's perception of the overall quality level. In addition, we calculate the fine-grained similarity, shown in~\cref{fig:4}, between AIGIs and the initial prompt to enhance the awareness of detailed quality. In the following, we present definite similarity measurements.

For ease of understanding, we briefly describe the notations in~\cref{tab:notation}.

\paragraph{Coarse-grained Similarity}
We can calculate the coarse-grained similarity $S^I_j$ between $F_{I_r}$ and $F_{{T_{ts}}_j}$ using \cref{eq:2} and convert it into a probability value by the Softmax function
\begin{equation}
	p^I_j=\frac{e^{S^I_j}}{\sum_{l=1}^{L}e^{S^I_l}}.
 	\label{eq:3}
\end{equation} 

Similarly, we can compute the coarse-grained similarity $S^P_j$ between patches and $F_{{T_{ts}}_j}$ and convert it into a probability value $p^p_j$ following \cref{eq:3}. Notably, we utilize average patch feature $\bar{F}_P$ of $\textit{\textbf{F}}_P$ for $S^P_j$ measurement.

Coarse-grained similarity evaluates the degree of alignment quality between an AIGI and quality-level descriptions, which provides an intuitive quality understanding. Additionally, we perform similarity measurements from both image and image patch perspectives, enhancing prediction accuracy and reliability.     

\paragraph{Fine-grained Similarity} 
Coarse-grained similarity computes the matching degree between an image and a sentence, focusing on general similarity. To account for finer details, we introduce the fine-grained similarity measurement between AIGI and its initial prompt. The initial prompt contains keywords used for image generation, by calculating the similarities between the image and these words, we can achieve a delicate awareness of image quality and content. To this end, we calculate the fine-grained similarity $S^I_k$ between $F_{I_r}$ and $F_{W_k}$ and average $K$ similarities, which is formalized as follows
\begin{equation} 
 w_I=\frac{1}{K}\sum_{k=1}^{K}\frac{F_{I_r}\odot F_{W_k}}{\left\|F_{I_r} \right\|\cdot \left\|F_{W_k} \right\|}.
  \label{eq:4}   
\end{equation}

Similarly, we can compute the mean fine-grained similarity $w_P$ between $\bar{F}_P$ and $F_{W_k}$ following \cref{eq:4}. Simultaneously considering the coarse-grained and fine-grained similarity, we achieve a holistic and detailed quality perception of AIGI, facilitating valid quality prediction.

\subsection{Quality Regression}
Here, we integrate and transform the above measurements into a quality score. First, $p^I_j$ and $p^P_j$ represent the probability that an AIGI belongs to the $j$-th quality level, thus, they are converted into quality scores with the following formula
\begin{equation}
  Q^*_{cg} =\frac{L}{L-1}\times(\sum_{j=1}^{L}j \times p^*_j-1),
    \label{eq:5}   
\end{equation}
where $*\in \left\{I,P \right\}$. 

\vspace{1em}

$w_I$  and $w_P$ indicate how well an AIGI matches the initial prompt, suggesting how good the detailed quality is. We convert them into quality scores with the following formula
\begin{equation}
	Q_{fg} = \frac{(w_I +w_P)}{2} \times L.
	\label{eq:6}   
\end{equation}
  
The final quality score of an AIGI is obtained by integrating $Q^*_{cg}$ and $Q_{fg}$, which is formalized as follows
\begin{equation}
     Q=\alpha\times Q^I_{cg}  + (1-\alpha)\times Q^P_{cg} + Q_{fg},
     \label{eq:7}
\end{equation}
where $\alpha$ is the weight used to balance $Q^I_{cg}$ and $Q^P_{cg}$.

Mean Absolute Error (MAE) is used as the loss function to fine-tune our model, which is shown as follows
\begin{equation}
	\mathcal{L} = \left|Q-Q' \right|,
	\label{eq:8}   
\end{equation}  
where $Q'$ is the subjective quality score.  

%% file: sec/4_experiments.tex
\section{Experiments}
\label{sec:Experiments}
This section briefly describes the AIGCIQA datasets, the metrics for performance evaluation, and the details of the implementation. Additionally, it provides a detailed analysis of the experimental results to offer insights into the proposed method.

\subsection{Datasets and Evaluation Criteria}
\paragraph{AIGCIQA Datasets} 
The proposed method is validated on two commonly used AIGCIQA datasets, including AGIQA-1K \cite{aigc1k}, and AGIQA-3K \cite{aigc3k}. The AGIQA-1K dataset contains 1080 images generated by two T2I models and each image is assigned a Mean Opinion Score (MOS) of quality. The AGIQA-3K dataset includes 2982 images generated by six T2I models, where MOS for both quality and alignment are provided. 

\paragraph{Evaluation Criteria} 
Spearman Rank Correlation Coefficient (SRCC) and Pearson Linear Correlation Coefficient (PLCC), defined as \cref{eq:srcc} and~\cref{eq:plcc}, are commonly employed as evaluation metrics for quality assessment. They measure the ranking ability and fitting ability of a prediction model respectively.
\vspace{1em}
\begin{equation}
	SRCC=1- \frac{6\sum_{m=1}^{M}r_{m}^{2}}{M\left ( M^{2}-1 \right )}
	\label{eq:srcc}
\end{equation}
\begin{equation}
    PLCC\!=\!\frac{\sum_{m=1}^{M}\!\left ( Q_{m}-\bar{Q} \right )\left ( Q'_{m}-\bar{Q'} \right )}{\sqrt{\sum_{m=1}^{M}\!\left ( Q_{m}\!-\!\bar{s} \right )^2}\sqrt{\sum_{m=1}^{M}\!\left ( Q'_{m}\!-\!\bar{Q'} \right )^2}}
	\label{eq:plcc}
\end{equation}
where $r_{m}$ is the rank difference of the subjective quality score $Q_{m}$ and the predicted score $Q'_{m}$.

\begin{table}
	\centering
	\setlength{\tabcolsep}{2mm}
	\begin{tabular}{ll|cc}
		\toprule
		Method & & SRCC &PLCC\\
		\midrule
		ResNet50~\cite{res50}  &\textit{CVPR'16} &0.6365&0.7323\\
		MGQA~\cite{MGIQ}  &\textit{VCIP'21} &0.6011 &0.6760\\
		CLIPIQA~\cite{CLIPiqa} &\textit{AAAI'23} &0.8227&0.8411 \\
	    IP-IQA~\cite{IP-IQA} &\textit{ICME'24} &0.8401 &\textcolor{Maroon}{0.8922} \\
		IPCE~\cite{IPCE} &\textit{CVPRW'24} &\textcolor{RoyalBlue}{0.8535} &0.8792\\
		MoE-AGIQA-v1~\cite{MoE-AGIQA} &\textit{CVPRW'24} &0.8530 &0.8877 \\
		MoE-AGIQA-v2~\cite{MoE-AGIQA} &\textit{CVPRW'24} &0.8501 &\textcolor{Maroon}{0.8922} \\
		\midrule
		TSP-MGS &-- &\textcolor{Maroon}{0.8567} &\textcolor{RoyalBlue}{0.8846}\\
		\bottomrule
	\end{tabular}
	\caption{Quantitative comparison on the AGIQA-1K. The top two results are highlighted in \textcolor{Maroon}{red} and \textcolor{RoyalBlue}{blue}, respectively.}
	\label{tab:agiqa1k}
\end{table}

\begin{table}
	\centering
	\setlength{\tabcolsep}{2mm}
	\begin{tabular}{@{}ll|cc@{}}
		\toprule
		\multicolumn{2}{c}{\multirow{2}{*}{Method}} &\multicolumn{2}{|c}{Perception}\\
		\cmidrule{3-4}
		& &SRCC &PLCC\\
		\midrule
		CNNIQA~\cite{deepiqa} &\textit{CVPR'14} &0.7478 &0.8469\\
		DBCNN~\cite{dbcnn}  &\textit{TCSVT'20} &0.8207&0.8759\\
		HyperIQA~\cite{hyperiqa}  &\textit{CVPR'20} &0.8355 &0.8903\\
		CLIPIQA~\cite{CLIPiqa} &\textit{AAAI'23} &0.8426&0.8053\\
		IP-IQA~\cite{IP-IQA} &\textit{ICME'24} &0.8634 &0.9116\\
		IPCE~\cite{IPCE} &\textit{CVPRW'24} &\textcolor{RoyalBlue}{0.8841} &0.9246\\
		MoE-AGIQA-v1~\cite{MoE-AGIQA} &\textit{CVPRW'24} &0.8758 &\textcolor{Maroon}{0.9294}\\
		MoE-AGIQA-v2~\cite{MoE-AGIQA} &\textit{CVPRW'24} &0.8746 &0.9014\\
		\midrule
		TSP-MGS &-- &\textcolor{Maroon}{0.8901} &\textcolor{RoyalBlue}{0.9270}\\
		\bottomrule
	\end{tabular}
	\caption{Performance comparison of perception quality prediction on the AGIQA-3K. The top two results are highlighted in \textcolor{Maroon}{red} and \textcolor{RoyalBlue}{blue}, respectively.}
	\label{tab:agiqa3k-p}
    \vspace{-1em}
\end{table}

\begin{table}
	\centering
	\setlength{\tabcolsep}{2.5mm}
	\begin{tabular}{@{}ll|cc@{}}
		\toprule
		\multicolumn{2}{c}{\multirow{2}{*}{Method}} &\multicolumn{2}{|c}{Alignment} \\
		\cmidrule{3-4}
		& &SRCC &PLCC \\
		\midrule
		CLIP~\cite{CLIP} &\textit{ICML'21} &0.5972&0.6839\\
		ImageReward~\cite{ImageReward}  &\textit{NIPS'23} &0.7298&0.7862\\
		HPS~\cite{HPS}  &\textit{ICCV'23} &0.6623&0.7008\\
		PickScore~\cite{PickScore}  &\textit{NIPS'23} &0.7320&0.7791\\
		StairReward~\cite{aigc3k}  &\textit{TCSVT'24} &0.7472&0.8529\\
		IPCE~\cite{IPCE} &\textit{CVPRW'24} &\textcolor{RoyalBlue}{0.7697} &\textcolor{RoyalBlue}{0.8725} \\
		\midrule
		TSP-MGS &-- &\textcolor{Maroon}{0.7734}&\textcolor{Maroon}{0.8773}\\
		\bottomrule
	\end{tabular}
	\caption{Performance comparison of alignment quality prediction on the AGIQA-3K. The top two results are highlighted in \textcolor{Maroon}{red} and \textcolor{RoyalBlue}{blue}, respectively.}
	\label{tab:agiqa3k-a}
\end{table}

\subsection{Implementation Details}
All experiments are performed on a PC equipped with an NVIDIA GeForce 4090 GPU, using PyTorch 1.12.0 and CUDA 11.3. We load the ViT-B/32 as the backbone of our method, where input images are with size $224\times224$. We employ the AdamW optimizer with a learning rate of 5e-6 and a weight decay of 5e-4. The model is trained for 20 epochs, with a cosine annealing learning rate scheduler applied to gradually reduce the learning rate every 5 epochs. Besides, we set the batch size to 16. 

To ensure the reproducibility of experiments and fairness in comparison, we adopt the dataset split strategy from the IPCE~\cite{IPCE} to divide each AIGCIQA dataset into training and testing sets in a 4:1 ratio. All experiments are conducted 10 times, and the average results are reported.                   
\subsection{Benchmark Results}
We compare the proposed method with existing deep learning (DL)-based AIGCIQA methods to highlight its superiority. 

\paragraph{AGIQA-1K} We compare our method with SOTA DL-based methods, including ResNet50~\cite{res50}, MGQA~\cite{MGIQ}, CLIPIQA~\cite{CLIPiqa}, IP-IQA~\cite{IP-IQA}, IPCE~\cite{IPCE}, and MoE-AGIQA-v1/v2~\cite{MoE-AGIQA}, on the AIGC-1K dataset. \cref{tab:agiqa1k} presents the SRCC and PLCC results, with the performance of the comparison methods sourced from existing work. It can be seen that multi-modal AIGCIQA methods~\cite{CLIPiqa,IP-IQA,IPCE,MoE-AGIQA} outperform single-modal ones~\cite{res50,MGIQ} by a considerable margin, indicating that prompts contribute to the AIGI quality perception. As a multi-modal method, TSP-MGS achieves a 0.8567 SRCC, surpassing other methods. In addition, it obtains a 0.8846 PLCC, the second-best result marginally lower than the best 0.8922.                  

\paragraph{AGIQA-3K} We compare our method with CNNIQA~\cite{deepiqa}, DBCNN~\cite{dbcnn}, HyperIQA~\cite{hyperiqa}, CLIPIQA~\cite{CLIPiqa}, IP-IQA~\cite{IP-IQA}, IPCE~\cite{IPCE}, and MoE-AGIQA-v1/v2~\cite{MoE-AGIQA} on the perception quality evaluation, and with VLM-based methods CLIP~\cite{CLIP}, ImageReward~\cite{ImageReward}, HPS~\cite{HPS}, PickScore~\cite{PickScore}, StairReward~\cite{aigc3k}, and IPCE~\cite{IPCE} on the alignment quality evaluation. The results are presented in \cref{tab:agiqa3k-p} and \cref{tab:agiqa3k-a}, respectively, where the performance of the comparison methods is sourced from existing work. 

As the results reported in \cref{tab:agiqa3k-p}, the methods~\cite{IP-IQA,IPCE,MoE-AGIQA} concentrated on degradation learning of AIGIs achieve better performance in perception quality predictions. Our method achieves the highest SRCC value of 0.8901 and the second-highest PLCC value of 0.9270, demonstrating its effectiveness in perceptual quality assessment. \cref{tab:agiqa3k-a} illustrates that all methods are weak in alignment quality prediction. However, our method shows superiority by achieving the best prediction performance with a 0.7734 SRCC and 0.8773 PLCC.                  

In conclusion, our method demonstrates robust performance in assessing perception and alignment quality on the AGIQA-1K and AGIQA-3K datasets, validating its effectiveness in addressing the complexities of AIGI. 

\begin{table}
	\centering
	\setlength{\tabcolsep}{1.6mm}
	\begin{tabular}{@{}c|cc|cc|cc@{}}
			\toprule
			\multirow{2}{*}{Task prompt} &\multirow{2}{*}{$I_r$} &\multirow{2}{*}{$\textit{\textbf{P}}$} &\multicolumn{2}{c|}{Perception} &\multicolumn{2}{c}{Alignment}\\
			\cmidrule{4-7}
			&& &SRCC &PLCC &SRCC &PLCC\\
			\cmidrule{1-7}
			\multirow{3}{*}{$\text{}{T^{ant}_{ts}}$} &\checkmark & &0.8766&0.9177 &0.7127&0.8304\\
			& &\checkmark &0.8891&0.9254 &0.7089&0.8401\\
			&\checkmark &\checkmark &0.8868 &0.9229 &0.7045&0.8341\\
			\cmidrule{1-7}
			
			\multirow{3}{*}{$\text{}{T^{adj}_{ts}}$} &\checkmark & &0.8780&0.9190 &0.7128&0.8406\\
			& &\checkmark &\textcolor{Maroon}{0.8901}&\textcolor{Maroon}{0.9270} &0.7129&0.8384\\
			&\checkmark &\checkmark &0.8893 &0.9266 &0.7115&0.8406\\
			\cmidrule{1-7}
			
			\multirow{3}{*}{$\text{}{T^{adv}_{ts}}$} &\checkmark & &0.8817&0.9214 &\textcolor{Maroon}{0.7734}&\textcolor{Maroon}{0.8773}\\
			& &\checkmark &0.8884&0.9245 &0.7544&0.8668 \\
			&\checkmark &\checkmark &0.8868 &0.9257 &0.7618&0.8729\\
			\bottomrule
		\end{tabular}
	\caption{The impact of different text prompts and image encoder inputs on AIGI quality assessment. The best results are highlighted in \textcolor{Maroon}{red}.}
	\label{tab:as-tp}
\end{table}

\subsection{Ablation Study}
To verify the effectiveness of each designed module of the proposed method, we perform ablation experiments on the AGIQA-3K dataset, which are detailed as follows.

\paragraph{Text Prompt} Here, we validate the effectiveness of task-specific prompts and initial prompts. First, we discuss the effects of different task-specific prompts $\text{}{T^{ant}_{ts}}$, $\text{}{T^{adj}_{ts}}$, and $\text{}{T^{adv}_{ts}}$. To be more persuasive, we compare the experimental results for each combination of task-specific prompts with the image encoder inputs $I_r$ and $\textit{\textbf{P}}$, as shown in \cref{tab:as-tp}. The best SRCC and PLCC are highlighted in \textcolor{Maroon}{red}. For perception quality evaluation, $\text{}{T^{adj}_{ts}}$, which emphasizes visual degradation levels as detailed in~\cref{sec:3.2}, boosts the model's prediction results. For alignment quality evaluation, $\text{}{T^{adv}_{ts}}$ further highlights the T2I correspondence degrees, leading to better predictions than the others. In addition, perception quality predictions based on image patches are generally better than those based on resized images, and alignment quality predictions using resized images are better than those using image patches. In summary, appropriately designing text descriptions for specific tasks enables the model to capture more relevant features. Moreover, an overall understanding of AIGIs benefits alignment quality evaluation, in which local detail perception promotes perception quality prediction. 

\begin{figure}[!t]
    \subfloat[]{
	\begin{minipage}[t]{.48\linewidth}
		\centering
		\includegraphics[width=1.\linewidth]{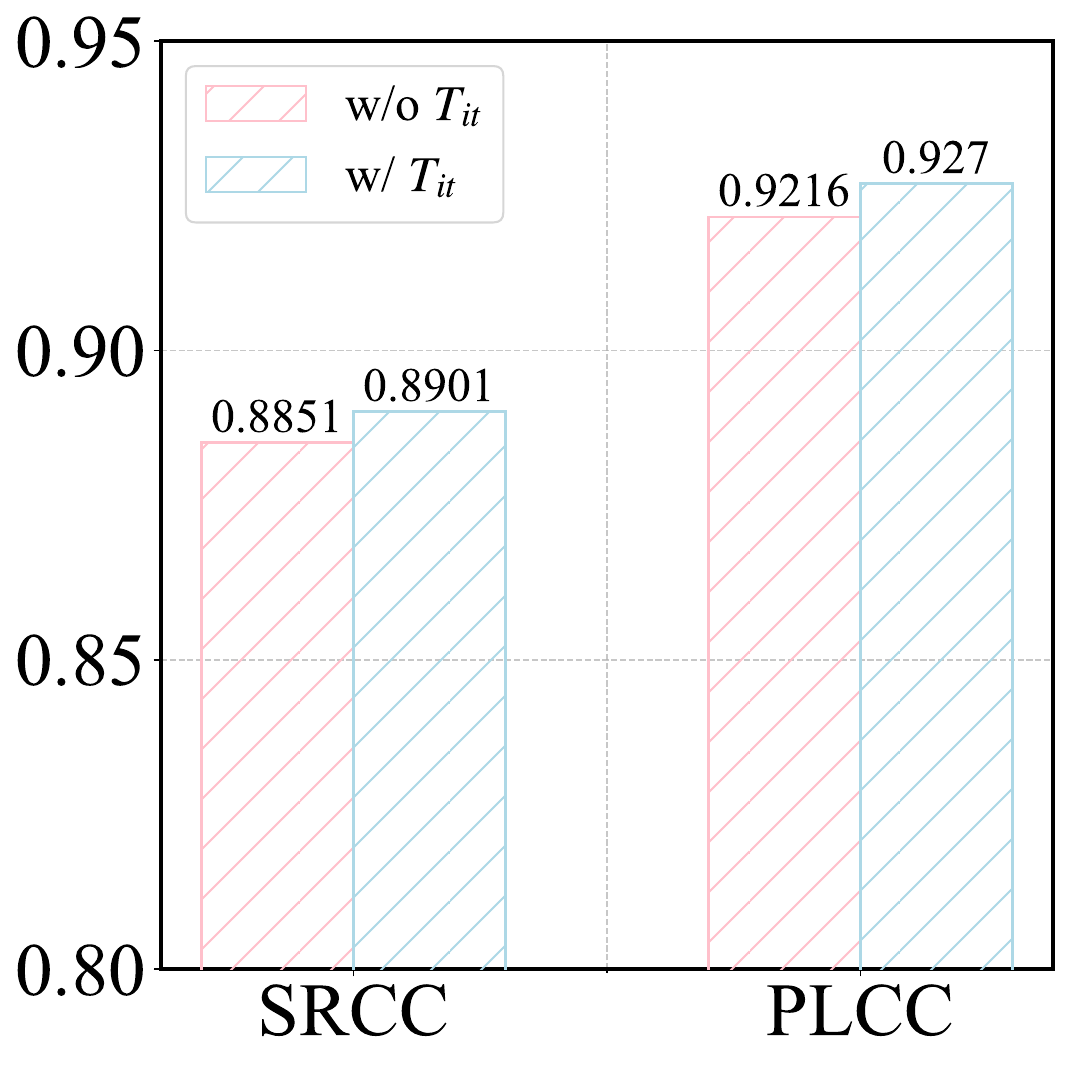}
	\end{minipage}
    \label{fig:5a}
    }
    \subfloat[]{
	\begin{minipage}[t]{.48\linewidth}
		\centering
		\includegraphics[width=1.\linewidth]{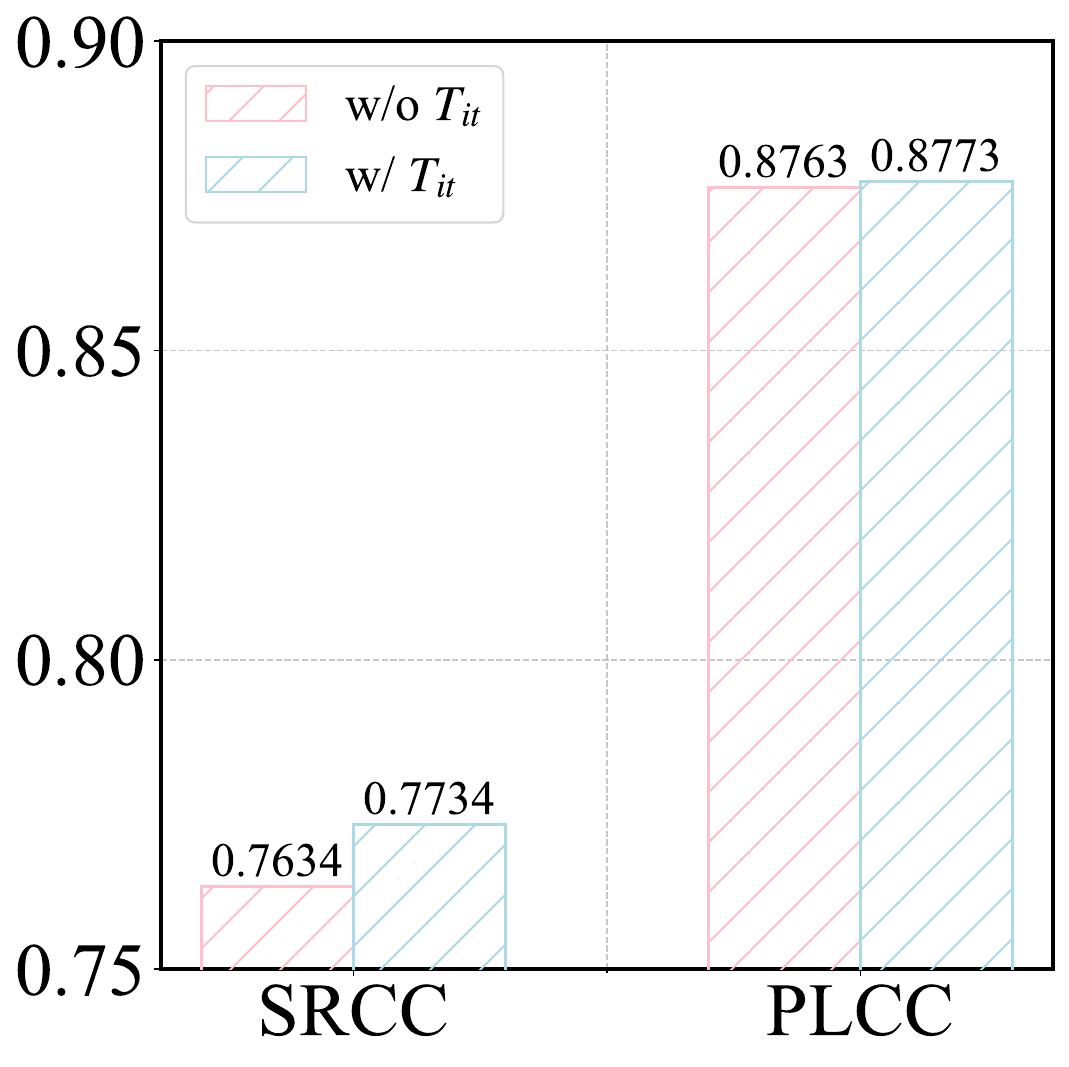}
	\end{minipage}
    \label{fig:5b}
    }
    \caption{The impact of the initial prompt on the (a) perception quality evaluation and (b) alignment quality evaluation, where $w/o$ means without and $w/$ represents with.}\label{fig:5}
\end{figure}
   
\begin{figure}[t]
    \subfloat[]{
	\begin{minipage}[t]{.48\linewidth}
		\centering
		\includegraphics[width=1.\linewidth]{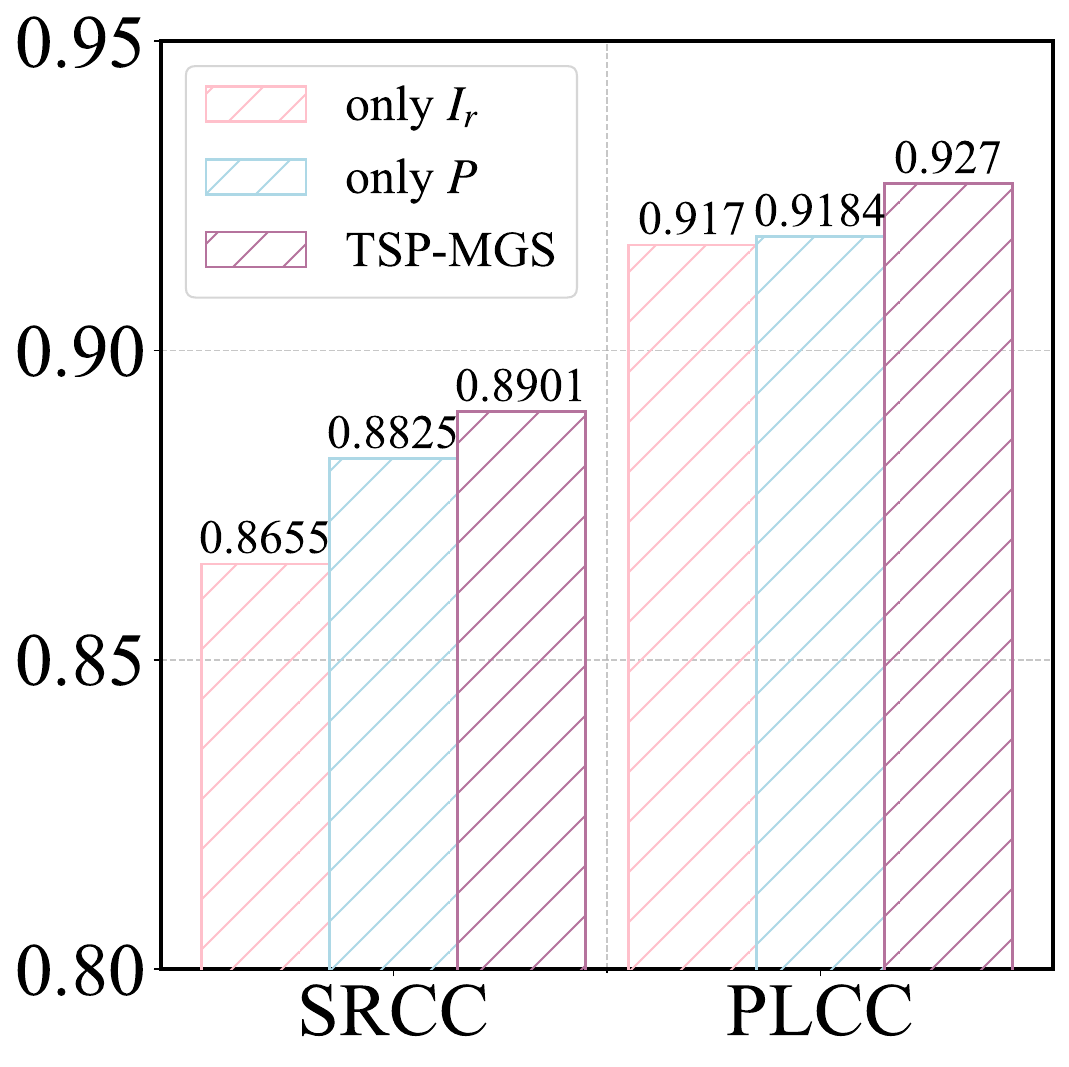}
	\end{minipage}
    \label{fig:6a}
    }
    \subfloat[]{
	\begin{minipage}[t]{.48\linewidth}
		\centering
		\includegraphics[width=1.\linewidth]{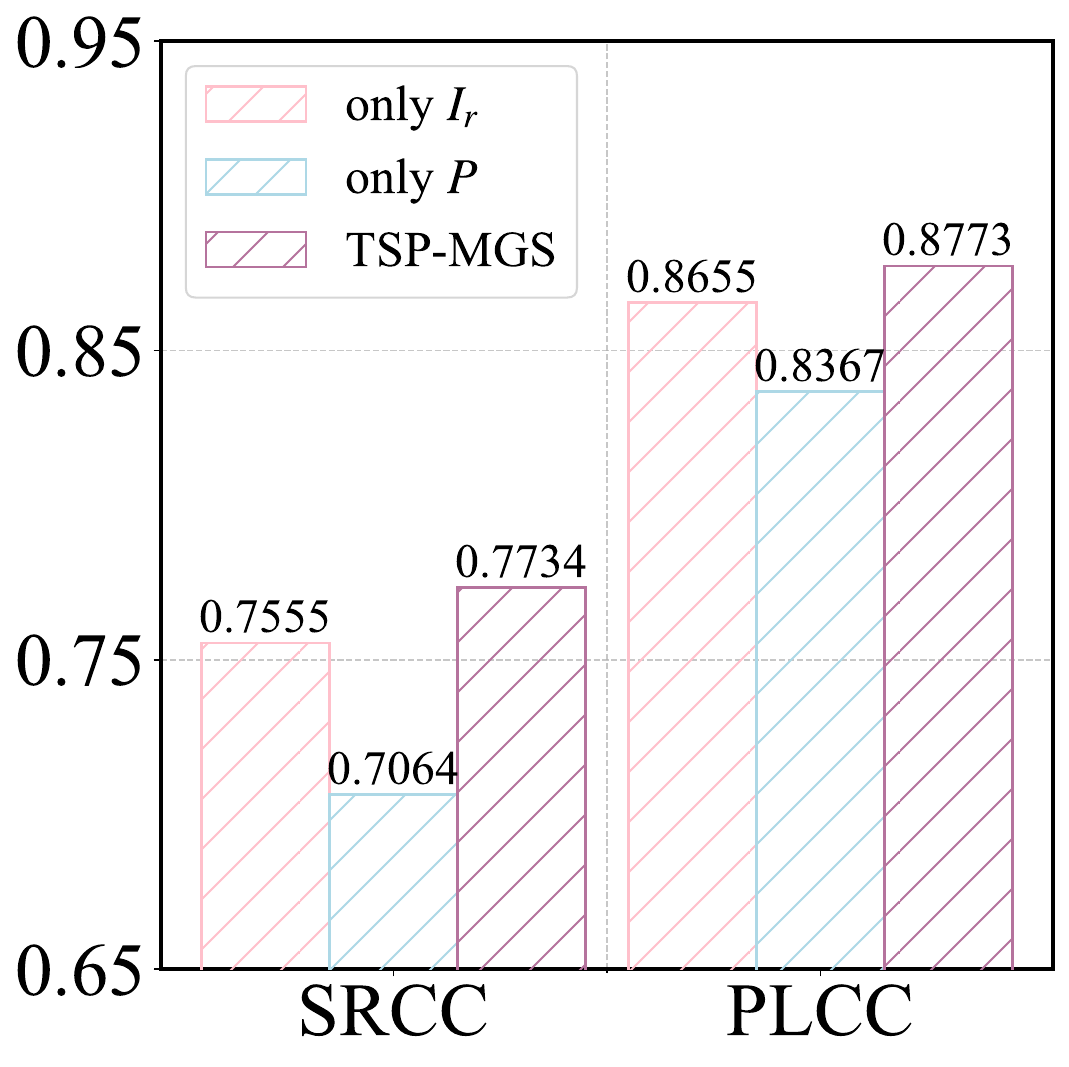}
	\end{minipage}
    \label{fig:6b}
    }
    \caption{The impact of the image encoder input on the (a) perception quality evaluation and (b) alignment quality evaluation.}  \label{fig:6}
\end{figure}

Additionally, we validate the effectiveness of the initial prompts, with results shown in \cref{fig:5}. It can be observed that excluding the initial prompts leads to reduced performance in both alignment quality and perception quality prediction. This indicates that enhancing the model's understanding of the relationship between AIGIs and prompt words can improve its ability to perceive image degradation.

\paragraph{Image Input} Following~\cite{IPCE}, we utilize the resized AIGI and its patches as inputs of the image encoder. Here, we analyze the model's performance using only the resized AIGI (Only $I_r$), and only the image patches (Only $P$). \cref{fig:6} illustrates the experimental results. Perception quality evaluation based only on $P$ surpasses that based only on $I_r$, demonstrating that more visual distortion details can be extracted from image patches. In contrast, alignment quality evaluation based only on $I_r$ outperforms that based only on $P$, underscoring the importance of alignment between text and overall image. Moreover, utilizing $I_r$ and $P$ yields rich image and patch feature combinations, significantly enhancing the model's performance in perception and alignment quality predictions. 

\begin{figure}[t]
    \subfloat[]{
	\begin{minipage}[t]{.48\linewidth}
		\centering
		\includegraphics[width=1.\linewidth]{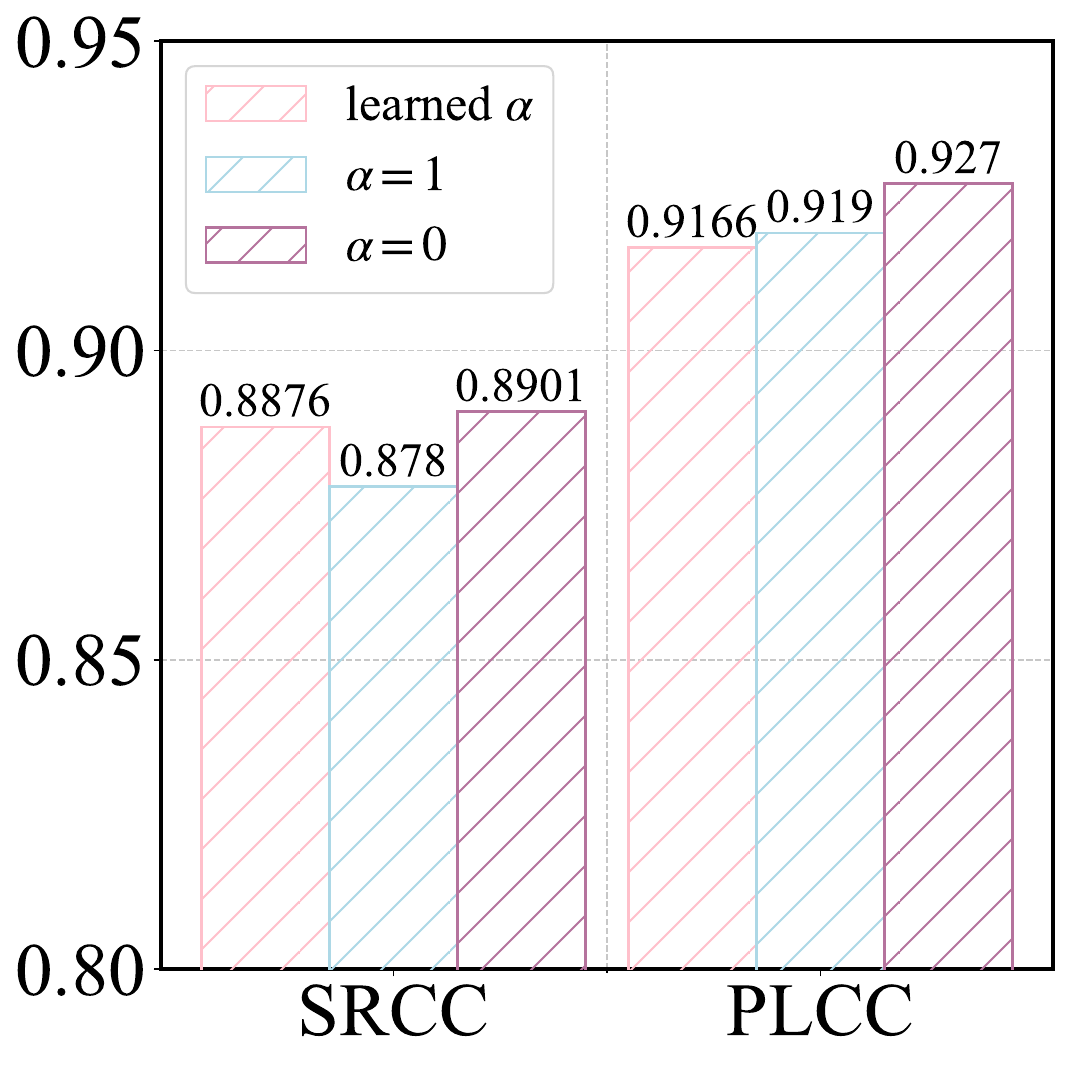}
	\end{minipage}
    \label{fig:8a}
    }
    \subfloat[]{
	\begin{minipage}[t]{.48\linewidth}
		\centering
		\includegraphics[width=1.\linewidth]{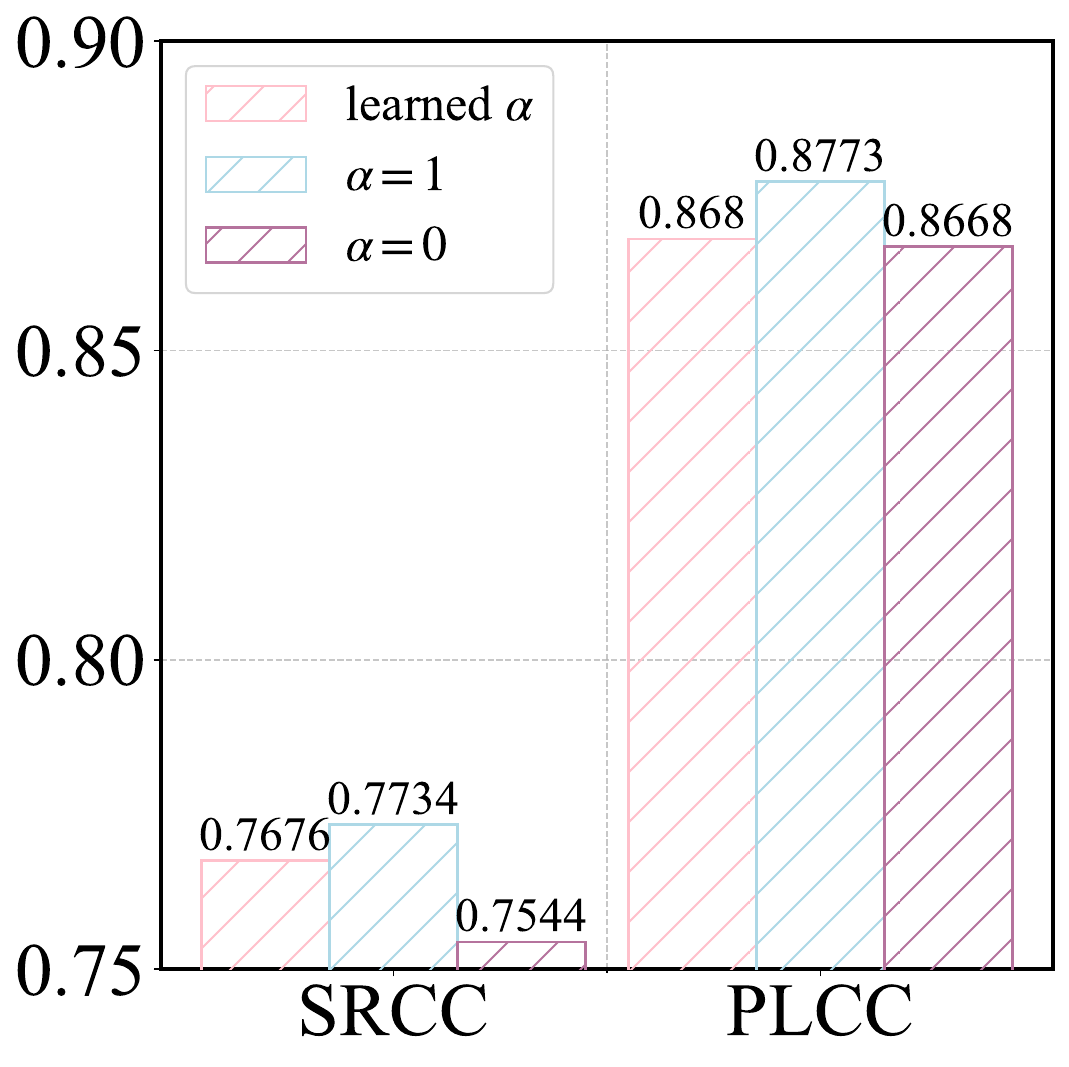}
	\end{minipage}
    \label{fig:8b}
    }
    \caption{The impact of $\alpha$ on the (a) perception quality evaluation and (b) alignment quality evaluation.}  \label{fig:8}
\end{figure}

\paragraph{Parameter $\alpha$} The parameter $\alpha$ mentioned in~\cref{eq:7} is used to balance $Q^I_{cg}$ and $Q^P_{cg}$. Here, we discuss its impact on the perception and alignment quality evaluation, including $\alpha=0$, $\alpha=1$, and $\alpha$ learned by the model. The results are shown in~\cref{fig:8}. For perceptual quality evaluation, the best prediction is achieved when setting $\alpha=0$, meaning that only coarse-grained similarity between image patches and perception-specific prompts is calculated. This highlights the importance of image details in perception quality prediction. For alignment quality evaluation, the optimal prediction is achieved when $\alpha=1$, where only the image is used in the coarse-grained similarity calculation. This emphasizes the importance of understanding the image content for alignment quality evaluation.     

\subsection{Visualization}
\begin{figure}[t]
	\centering
    \subfloat[]{
	\begin{minipage}[t]{.32\linewidth}
		\centering
		\includegraphics[width=.96\linewidth]{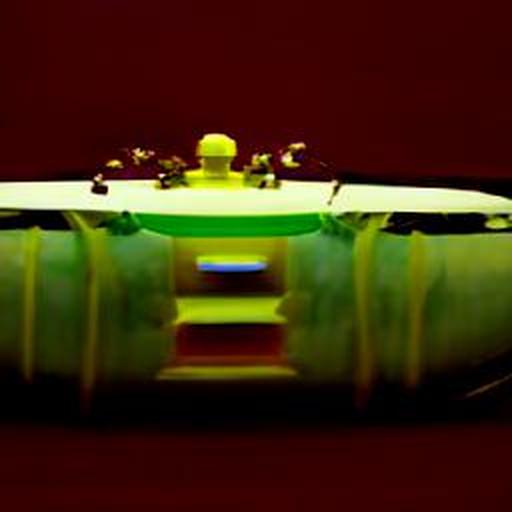}
		\captionsetup{justification=raggedright, singlelinecheck=false, font=footnotesize} 
		\caption*{
        \textcolor{RawSienna}{Alignment: 0.8149}
        \textcolor{RoyalBlue}{Predicted: 0.8242}
        ------------------------
        \textcolor{RawSienna}{Perception: 1.2992}
        \textcolor{RoyalBlue}{Predicted: 1.2314}
        }
	\end{minipage}
	\begin{minipage}[t]{.32\linewidth}
		\includegraphics[width=.96\linewidth]{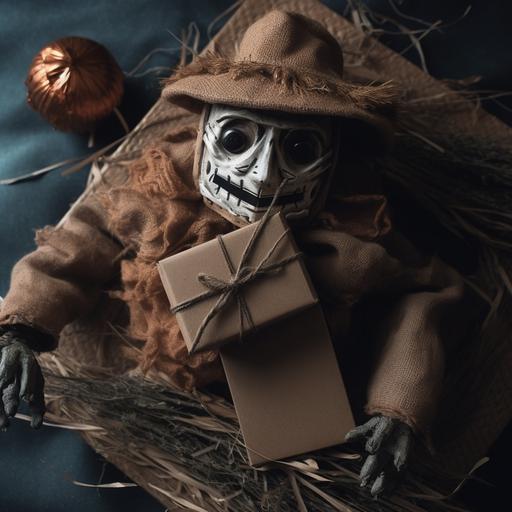}
		\captionsetup{justification=raggedright, singlelinecheck=false, font=footnotesize} 
		\caption*{        
        \textcolor{RawSienna}{Alignment: 3.3075}
        \textcolor{RoyalBlue}{Predicted: 3.2813}
        ------------------------
        \textcolor{RawSienna}{Perception: 3.3619}
        \textcolor{RoyalBlue}{Predicted: 3.4082}}
	\end{minipage}
	\begin{minipage}[t]{.32\linewidth}
		\includegraphics[width=.96\linewidth]{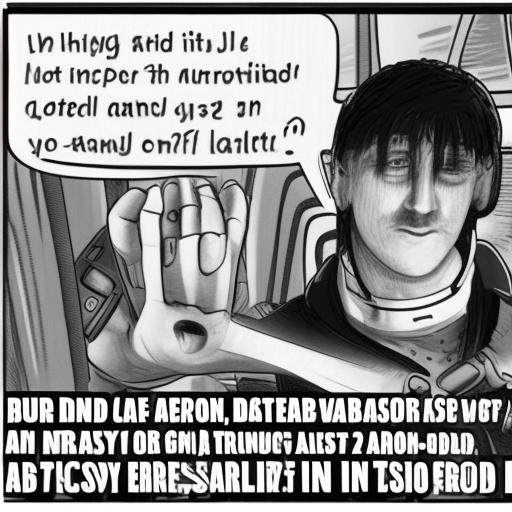}
		\captionsetup{justification=raggedright, singlelinecheck=false, font=footnotesize} 
		\caption*{
        \textcolor{RawSienna}{Alignment: 1.7343}
        \textcolor{RoyalBlue}{Predicted: 1.734}
        ------------------------
        \textcolor{RawSienna}{Perception: 2.2232}
        \textcolor{RoyalBlue}{Predicted: 2.2304}
        }
	\end{minipage}
    \label{fig:7a}
    }

    \subfloat[]{
	\begin{minipage}[t]{.32\linewidth}
		\centering
		\includegraphics[width=.96\linewidth]{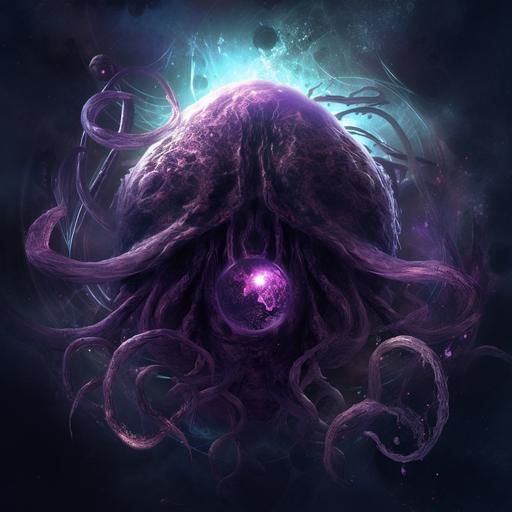}
		\captionsetup{justification=raggedright, singlelinecheck=false, font=footnotesize} 
	\caption*{
            \textcolor{RawSienna}{Alignment: 3.3645}
            \textcolor{RoyalBlue}{Predicted: 3.3691}
            ------------------------
            \textcolor{RawSienna}{Perception: 2.7423}
            \textcolor{RoyalBlue}{Predicted: 3.3965} $\times$
        }
    \end{minipage}
        \begin{minipage}[t]{.32\linewidth}
	\includegraphics[width=.96\linewidth]{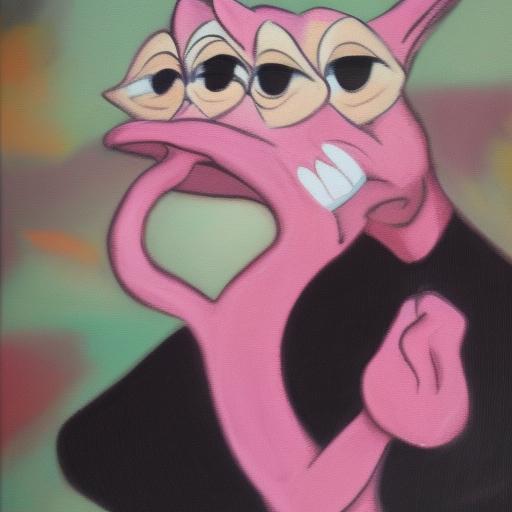}
		\captionsetup{justification=raggedright, singlelinecheck=false, font=footnotesize} 
		\caption*{
        \textcolor{RawSienna}{Alignment: 1.1654}
        \textcolor{RoyalBlue}{Predicted: 2.3027} $\times$
        ------------------------
        \textcolor{RawSienna}{Perception: 1.5547}
        \textcolor{RoyalBlue}{Predicted: 1.5576}
        }
	\end{minipage}
	\begin{minipage}[t]{.32\linewidth}
		\includegraphics[width=.96\linewidth]{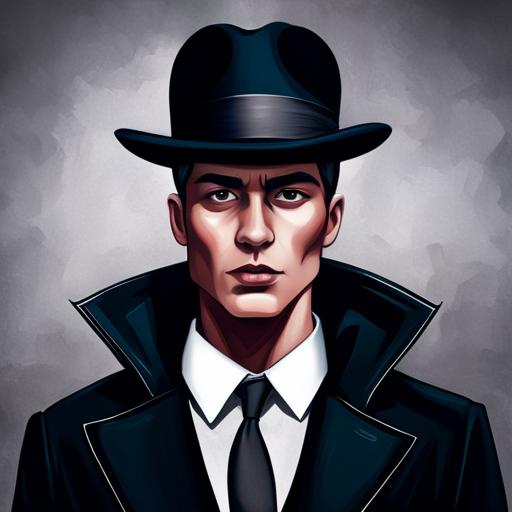}
		\captionsetup{justification=raggedright, singlelinecheck=false, font=footnotesize} 
	\caption*{
            \textcolor{RawSienna}{Alignment: 0.5641}
            \textcolor{RoyalBlue}{Predicted: 2.1758} $\times$
            ------------------------
            \textcolor{RawSienna}{Perception: 3.7276}
            \textcolor{RoyalBlue}{Predicted: 3.6816}
        }
	\end{minipage}
    \label{fig:7b}
    }
	 \caption{Illustration of some AIGIs with subjective score (marked in \textcolor{RawSienna}{red}) and prediction (marked in \textcolor{RoyalBlue}{blue}) by TSP-MGS. $\times$ is used to signal poor prediction. }\label{fig:7}
     \vspace{2em}
\end{figure}

To provide an intuitive observation, we present several AIGIs with subjective and predicted scores, shown in~\cref{fig:7}. ~\cref{fig:7a} shows the accurately predicted samples covering a rich generated image content. The proposed TSP-MGS achieves precise perception and alignment quality predictions on these samples, demonstrating its effectiveness in handling complex AIGI quality evaluation. ~\cref{fig:7b} shows the samples with failed predictions. These samples either have high alignment quality predictions but low perception predictions, or the reverse. The causes of the failures may be (1) AIGIs exhibiting rich textures but poor color, lighting, etc., which are often difficult for the model to capture. This leads to a deviation from a subjective score in the perception prediction, even when the predicted alignment quality is accurate. (2) Bias in the model's understanding of the prompt or ambiguity arising from multiple meanings of words, failing alignment quality prediction. In summary, improving the model's perception of non-structural distortions and promoting its correct understanding of prompts are crucial to improving the perception and alignment quality evaluation performance.         

%% file: sec/5_conclusion.tex
\section{Conclusion}
\label{sec:Conclusions}

This paper introduces a novel quality assessment method for AIGIs, named TSP-MGS. It employs the CLIP model to measure the similarity between AIGI and the constructed prompts and transforms the similarity into a quality score. In summary, TSP-MGS has two innovations: (1) Considering the inconsistency between alignment and perception quality evaluation, task-specific prompts are constructed to guide each task learning, achieving holistic quality awareness. Moreover, the initial prompt used to generate the image is also introduced for a detailed quality perception. (2) Multi-granularity similarity between image and prompts is measured and integrated to enhance the model's capabilities in intuitive quality understanding and detailed quality awareness. Experimental results on the AGIQA-1K and AGIQA-3K datasets validate the effectiveness of TSP-MGS in both alignment and perception quality assessment.

%% file: main.bbl
\begin{thebibliography}{72}
\providecommand{\natexlab}[1]{#1}
\providecommand{\url}[1]{\texttt{#1}}
\expandafter\ifx\csname urlstyle\endcsname\relax
  \providecommand{\doi}[1]{doi: #1}\else
  \providecommand{\doi}{doi: \begingroup \urlstyle{rm}\Url}\fi

\bibitem[Cheon et~al.(2021)Cheon, Yoon, Kang, and Lee]{IQT}
Manri Cheon, Sung-Jun Yoon, Byungyeon Kang, and Junwoo Lee.
\newblock Perceptual image quality assessment with transformers.
\newblock In \emph{CVPRW}, pages 433--442, 2021.

\bibitem[Deng et~al.(2023)Deng, Shi, Huang, Li, Xu, Han, Kwok, Zhao, Zhang, and
  Liang]{GrowCLIP}
Xinchi Deng, Han Shi, Runhui Huang, Changlin Li, Hang Xu, Jianhua Han, James
  Kwok, Shen Zhao, Wei Zhang, and Xiaodan Liang.
\newblock Growclip: Data-aware automatic model growing for large-scale
  contrastive language-image pre-training.
\newblock In \emph{ICCV}, pages 22121--22132, 2023.

\bibitem[Ding et~al.(2022)Ding, Ma, Wang, and Simoncelli]{DISTS}
Keyan Ding, Kede Ma, Shiqi Wang, and Eero~P. Simoncelli.
\newblock Image quality assessment: Unifying structure and texture similarity.
\newblock \emph{IEEE TPAMI}, 44\penalty0 (5):\penalty0 2567--2581, 2022.

\bibitem[Ding et~al.(2021)Ding, Yang, Hong, Zheng, Zhou, Yin, Lin, Zou, Shao,
  Yang, et~al.]{Cogview}
Ming Ding, Zhuoyi Yang, Wenyi Hong, Wendi Zheng, Chang Zhou, Da Yin, Junyang
  Lin, Xu Zou, Zhou Shao, Hongxia Yang, et~al.
\newblock Cogview: Mastering text-to-image generation via transformers.
\newblock \emph{NIPS}, 34:\penalty0 19822--19835, 2021.

\bibitem[Fang et~al.(2024)Fang, Wang, Lv, and Yan]{PCQA}
Xi Fang, Weigang Wang, Xiaoxin Lv, and Jun Yan.
\newblock Pcqa: A strong baseline for aigc quality assessment based on prompt
  condition.
\newblock In \emph{CVPRW}, pages 6167--6176, 2024.

\bibitem[Fang et~al.(2020)Fang, Zhu, Zeng, Ma, and Wang]{spaq}
Yuming Fang, Hanwei Zhu, Yan Zeng, Kede Ma, and Zhou Wang.
\newblock Perceptual quality assessment of smartphone photography.
\newblock In \emph{CVPR}, pages 3674--3683, 2020.

\bibitem[Fu et~al.(2024)Fu, Zhou, Jiang, Liu, and Zhai]{CLIP-AGIQA}
Jun Fu, Wei Zhou, Qiuping Jiang, Hantao Liu, and Guangtao Zhai.
\newblock Vision-language consistency guided multi-modal prompt learning for
  blind ai generated image quality assessment.
\newblock \emph{IEEE Signal Processing Letters}, 31:\penalty0 1820--1824, 2024.

\bibitem[Gao et~al.(2013)Gao, Gao, Tao, and Li]{MKL}
Xinbo Gao, Fei Gao, Dacheng Tao, and Xuelong Li.
\newblock Universal blind image quality assessment metrics via natural scene
  statistics and multiple kernel learning.
\newblock \emph{IEEE Transactions on Neural Networks and Learning Systems},
  24\penalty0 (12):\penalty0 2013--2026, 2013.

\bibitem[Gao et~al.(2024)Gao, Min, Zhu, Zhang, and Zhai]{fuzzy}
Yixuan Gao, Xiongkuo Min, Yucheng Zhu, Xiao-Ping Zhang, and Guangtao Zhai.
\newblock Blind image quality assessment: A fuzzy neural network for opinion
  score distribution prediction.
\newblock \emph{IEEE Transactions on Circuits and Systems for Video
  Technology}, 34\penalty0 (3):\penalty0 1641--1655, 2024.

\bibitem[Ghadiyaram and Bovik(2016)]{livec}
Deepti Ghadiyaram and Alan~C. Bovik.
\newblock Massive online crowdsourced study of subjective and objective picture
  quality.
\newblock \emph{IEEE TIP}, 25\penalty0 (1):\penalty0 372--387, 2016.

\bibitem[Golestaneh et~al.(2022)Golestaneh, Dadsetan, and Kitani]{tres}
S.~Alireza Golestaneh, Saba Dadsetan, and Kris~M. Kitani.
\newblock No-reference image quality assessment via transformers, relative
  ranking, and self-consistency.
\newblock pages 3989--3999, 2022.

\bibitem[He et~al.(2016)He, Zhang, Ren, and Sun]{res50}
Kaiming He, Xiangyu Zhang, Shaoqing Ren, and Jian Sun.
\newblock Deep residual learning for image recognition.
\newblock In \emph{CVPR}, pages 770--778, 2016.

\bibitem[Hosu et~al.(2020)Hosu, Lin, Sziranyi, and Saupe]{koniq-10k}
Vlad Hosu, Hanhe Lin, Tamas Sziranyi, and Dietmar Saupe.
\newblock Koniq-10k: An ecologically valid database for deep learning of blind
  image quality assessment.
\newblock \emph{IEEE TIP}, 29:\penalty0 4041--4056, 2020.

\bibitem[Jia et~al.(2021)Jia, Yang, Xia, Chen, Parekh, Pham, Le, Sung, Li, and
  Duerig]{ALIGN}
Chao Jia, Yinfei Yang, Ye Xia, Yi-Ting Chen, Zarana Parekh, Hieu Pham, Quoc Le,
  Yun-Hsuan Sung, Zhen Li, and Tom Duerig.
\newblock Scaling up visual and vision-language representation learning with
  noisy text supervision.
\newblock In \emph{International conference on machine learning}, pages
  4904--4916. PMLR, 2021.

\bibitem[Kang et~al.(2014)Kang, Ye, Li, and Doermann]{deepiqa}
Le Kang, Peng Ye, Yi Li, and David Doermann.
\newblock Convolutional neural networks for no-reference image quality
  assessment.
\newblock In \emph{CVPR}, pages 1733--1740, 2014.

\bibitem[Kirstain et~al.(2024)Kirstain, Polyak, Singer, Matiana, Penna, and
  Levy]{PickScore}
Yuval Kirstain, Adam Polyak, Uriel Singer, Shahbuland Matiana, Joe Penna, and
  Omer Levy.
\newblock Pick-a-pic: an open dataset of user preferences for text-to-image
  generation.
\newblock 2024.

\bibitem[Kumari et~al.(2023)Kumari, Zhang, Zhang, Shechtman, and
  Zhu]{Diffusion02}
Nupur Kumari, Bingliang Zhang, Richard Zhang, Eli Shechtman, and Jun-Yan Zhu.
\newblock Multi-concept customization of text-to-image diffusion.
\newblock In \emph{CVPR}, pages 1931--1941, 2023.

\bibitem[Larson and Chandler(2010)]{csiq}
Eric~Cooper Larson and Damon~Michael Chandler.
\newblock Most apparent distortion: full-reference image quality assessment and
  the role of strategy.
\newblock \emph{J. Electron. Imaging}, 19\penalty0 (1):\penalty0 011006, 2010.

\bibitem[Li et~al.(2024)Li, Zhang, Wu, Sun, Min, Liu, Zhai, and Lin]{aigc3k}
Chunyi Li, Zicheng Zhang, Haoning Wu, Wei Sun, Xiongkuo Min, Xiaohong Liu,
  Guangtao Zhai, and Weisi Lin.
\newblock Agiqa-3k: An open database for ai-generated image quality assessment.
\newblock \emph{IEEE Transactions on Circuits and Systems for Video
  Technology}, 34\penalty0 (8):\penalty0 6833--6846, 2024.

\bibitem[Li et~al.(2022)Li, Zhang, Zhang, Yang, Li, Zhong, Wang, Yuan, Zhang,
  Hwang, Chang, and Gao]{GLIP}
Liunian~Harold Li, Pengchuan Zhang, Haotian Zhang, Jianwei Yang, Chunyuan Li,
  Yiwu Zhong, Lijuan Wang, Lu Yuan, Lei Zhang, Jenq-Neng Hwang, Kai-Wei Chang,
  and Jianfeng Gao.
\newblock Grounded language-image pre-training.
\newblock In \emph{CVPR}, pages 10955--10965, 2022.

\bibitem[Lin et~al.(2019)Lin, Hosu, and Saupe]{kadid-10k}
Hanhe Lin, Vlad Hosu, and Dietmar Saupe.
\newblock Kadid-10k: A large-scale artificially distorted iqa database.
\newblock In \emph{Int. Conf. Qual. Multimedia Exp.}, pages 1--3, 2019.

\bibitem[Lin and Wang(2018)]{Hallucinated-IQA}
Kwan-Yee Lin and Guanxiang Wang.
\newblock Hallucinated-iqa: No-reference image quality assessment via
  adversarial learning.
\newblock In \emph{CVPR}, pages 732--741, 2018.

\bibitem[Liu et~al.(2017)Liu, Van De~Weijer, and Bagdanov]{rankiqa}
Xialei Liu, Joost Van De~Weijer, and Andrew~D. Bagdanov.
\newblock Rankiqa: Learning from rankings for no-reference image quality
  assessment.
\newblock In \emph{ICCV}, pages 1040--1049, 2017.

\bibitem[Luo et~al.(2023)Luo, Bao, Wu, He, and Li]{Segclip}
Huaishao Luo, Junwei Bao, Youzheng Wu, Xiaodong He, and Tianrui Li.
\newblock Segclip: Patch aggregation with learnable centers for open-vocabulary
  semantic segmentation.
\newblock In \emph{International Conference on Machine Learning}, pages
  23033--23044. PMLR, 2023.

\bibitem[Mittal et~al.(2012)Mittal, Moorthy, and Bovik]{BRISQUE}
Anish Mittal, Anush~Krishna Moorthy, and Alan~Conrad Bovik.
\newblock No-reference image quality assessment in the spatial domain.
\newblock \emph{IEEE TIP}, 21\penalty0 (12):\penalty0 4695--4708, 2012.

\bibitem[Moorthy and Bovik(2011)]{DIIVINE}
Anush~Krishna Moorthy and Alan~Conrad Bovik.
\newblock Blind image quality assessment: From natural scene statistics to
  perceptual quality.
\newblock \emph{IEEE Transactions on Image Processing}, 20\penalty0
  (12):\penalty0 3350--3364, 2011.

\bibitem[Pan et~al.(2022)Pan, Zhang, Lei, Fang, Shao, Ling, and Kwong]{dacnn}
Zhaoqing Pan, Hao Zhang, Jianjun Lei, Yuming Fang, Xiao Shao, Nam Ling, and Sam
  Kwong.
\newblock Dacnn: Blind image quality assessment via a distortion-aware
  convolutional neural network.
\newblock \emph{IEEE Transactions on Circuits and Systems for Video
  Technology}, 32\penalty0 (11):\penalty0 7518--7531, 2022.

\bibitem[Peng et~al.(2024)Peng, Fu, Ming, Wang, Ma, He, Dou, and Chen]{IPCE}
Fei Peng, Huiyuan Fu, Anlong Ming, Chuanming Wang, Huadong Ma, Shuai He, Zifei
  Dou, and Shu Chen.
\newblock Aigc image quality assessment via image-prompt correspondence.
\newblock In \emph{CVPRW}, pages 6432--6441, 2024.

\bibitem[Ponomarenko et~al.(2013)Ponomarenko, Ieremeiev, Lukin, Egiazarian,
  Jin, Astola, Vozel, Chehdi, Carli, and Battisti]{tid2013}
Nikolay Ponomarenko, Oleg Ieremeiev, Vladimir Lukin, Karen Egiazarian, Lina
  Jin, Jaakko Astola, Benoit Vozel, Kacem Chehdi, Marco Carli, and Federica
  Battisti.
\newblock Color image database tid2013: Peculiarities and preliminary results.
\newblock In \emph{EUVIP}, pages 106--111, 2013.

\bibitem[Qu et~al.(2024)Qu, Li, and Gao]{IP-IQA}
Bowen Qu, Haohui Li, and Wei Gao.
\newblock Bringing textual prompt to ai-generated image quality assessment.
\newblock In \emph{ICME}, pages 1--6, 2024.

\bibitem[Radford et~al.(2021)Radford, Kim, Hallacy, Ramesh, Goh, Agarwal,
  Sastry, Askell, Mishkin, Clark, et~al.]{CLIP}
Alec Radford, Jong~Wook Kim, Chris Hallacy, Aditya Ramesh, Gabriel Goh,
  Sandhini Agarwal, Girish Sastry, Amanda Askell, Pamela Mishkin, Jack Clark,
  et~al.
\newblock Learning transferable visual models from natural language
  supervision.
\newblock In \emph{International conference on machine learning}, pages
  8748--8763. PMLR, 2021.

\bibitem[Ramesh et~al.(2021)Ramesh, Pavlov, Goh, Gray, Voss, Radford, Chen, and
  Sutskever]{regression02}
Aditya Ramesh, Mikhail Pavlov, Gabriel Goh, Scott Gray, Chelsea Voss, Alec
  Radford, Mark Chen, and Ilya Sutskever.
\newblock Zero-shot text-to-image generation.
\newblock In \emph{ICML. PMLR}, pages 8821--8831, 2021.

\bibitem[Saad et~al.(2012)Saad, Bovik, and Charrier]{BLIINDS-II}
Michele~A. Saad, Alan~C. Bovik, and Christophe Charrier.
\newblock Blind image quality assessment: A natural scene statistics approach
  in the dct domain.
\newblock \emph{IEEE Transactions on Image Processing}, 21\penalty0
  (8):\penalty0 3339--3352, 2012.

\bibitem[Saha et~al.(2023)Saha, Mishra, and Bovik]{re-iqa}
Avinab Saha, Sandeep Mishra, and Alan~C. Bovik.
\newblock Re-iqa: Unsupervised learning for image quality assessment in the
  wild.
\newblock In \emph{CVPR}, pages 5846--5855, 2023.

\bibitem[Schuhmann et~al.(2022)Schuhmann, Beaumont, Vencu, Gordon, Wightman,
  Cherti, Coombes, Katta, Mullis, Wortsman, et~al.]{Laion-5b}
Christoph Schuhmann, Romain Beaumont, Richard Vencu, Cade Gordon, Ross
  Wightman, Mehdi Cherti, Theo Coombes, Aarush Katta, Clayton Mullis, Mitchell
  Wortsman, et~al.
\newblock Laion-5b: An open large-scale dataset for training next generation
  image-text models.
\newblock \emph{Advances in Neural Information Processing Systems},
  35:\penalty0 25278--25294, 2022.

\bibitem[Sheikh et~al.(2006)Sheikh, Sabir, and Bovik]{live}
Hamid~R Sheikh, Muhammad~F Sabir, and Alan~C. Bovik.
\newblock A statistical evaluation of recent full reference image quality
  assessment algorithms.
\newblock \emph{TIP}, 15\penalty0 (11):\penalty0 3440--3451, 2006.

\bibitem[Shi et~al.(2024)Shi, Gao, and Smolic]{TempQT}
Jinsong Shi, Pan Gao, and Aljosa Smolic.
\newblock Blind image quality assessment via transformer predicted error map
  and perceptual quality token.
\newblock \emph{IEEE Transactions on Multimedia}, 26:\penalty0 4641--4651,
  2024.

\bibitem[Shin et~al.(2024)Shin, Lee, and Kim]{QCN}
Nyeong-Ho Shin, Seon-Ho Lee, and Chang-Su Kim.
\newblock Blind image quality assessment based on geometric order learning.
\newblock In \emph{CVPR}, pages 12799--12808, 2024.

\bibitem[Singh et~al.(2022)Singh, Hu, Goswami, Couairon, Galuba, Rohrbach, and
  Kiela]{FLAVA}
Amanpreet Singh, Ronghang Hu, Vedanuj Goswami, Guillaume Couairon, Wojciech
  Galuba, Marcus Rohrbach, and Douwe Kiela.
\newblock Flava: A foundational language and vision alignment model.
\newblock In \emph{CVPR}, pages 15617--15629, 2022.

\bibitem[Su et~al.(2020)Su, Yan, Zhu, Zhang, Ge, Sun, and Zhang]{hyperiqa}
Shaolin Su, Qingsen Yan, Yu Zhu, Cheng Zhang, Xin Ge, Jinqiu Sun, and Yanning
  Zhang.
\newblock Blindly assess image quality in the wild guided by a self-adaptive
  hyper network.
\newblock In \emph{CVPR}, pages 3664--3673, 2020.

\bibitem[Sun et~al.(2022)Sun, Duan, Min, Chen, and Zhai]{StairIQA}
Wei Sun, Huiyu Duan, Xiongkuo Min, Li Chen, and Guangtao Zhai.
\newblock Blind quality assessment for in-the-wild images via hierarchical
  feature fusion strategy.
\newblock In \emph{IEEE International Symposium on Broadband Multimedia Systems
  and Broadcasting}, pages 01--06, 2022.

\bibitem[Wang et~al.(2022)Wang, Fan, Hou, Xu, Li, Lu, and Fu]{wang}
Jing Wang, Haotian Fan, Xiaoxia Hou, Yitian Xu, Tao Li, Xuechao Lu, and Lean
  Fu.
\newblock Mstriq: No reference image quality assessment based on swin
  transformer with multi-stage fusion.
\newblock In \emph{CVPRW}, pages 1268--1277, 2022.

\bibitem[Wang et~al.(2023)Wang, Chan, and Loy]{CLIPiqa}
Jianyi Wang, Kelvin~C.K. Chan, and Chen~Change Loy.
\newblock Exploring clip for assessing the look and feel of images.
\newblock In \emph{AAAI}, pages 2555--2563, 2023.

\bibitem[Wang et~al.(2024)Wang, Duan, Liu, Chen, Min, and Zhai]{AIGCIQA2023}
Jiarui Wang, Huiyu Duan, Jing Liu, Shi Chen, Xiongkuo Min, and Guangtao Zhai.
\newblock Aigciqa2023: A large-scale image quality assessment database for ai
  generated images: From the perspectives of quality, authenticity and
  correspondence.
\newblock In \emph{CAAI International Conference on Artificial Intelligence},
  pages 46--57, 2024.

\bibitem[Wang et~al.(2021)Wang, Sun, Min, Lu, Zhang, and Zhai]{MGIQ}
Tao Wang, Wei Sun, Xiongkuo Min, Wei Lu, Zicheng Zhang, and Guangtao Zhai.
\newblock A multi-dimensional aesthetic quality assessment model for mobile
  game images.
\newblock In \emph{International Conference on Visual Communications and Image
  Processing}, pages 1--5, 2021.

\bibitem[Wang and Ma(2022)]{gmad}
Zhihua Wang and Kede Ma.
\newblock Active fine-tuning from gmad examples improves blind image quality
  assessment.
\newblock \emph{IEEE TPAMI}, 44\penalty0 (9):\penalty0 4577--4590, 2022.

\bibitem[Wu et~al.(2024{\natexlab{a}})Wu, Zhang, Zhang, Chen, Liao, Wang, Li,
  Sun, Yan, Zhai, and Lin]{qbench}
Haoning Wu, Zicheng Zhang, Erli Zhang, Chaofeng Chen, Liang Liao, Annan Wang,
  Chunyi Li, Wenxiu Sun, Qiong Yan, Guangtao Zhai, and Weisi Lin.
\newblock Q-bench: A benchmark for general-purpose foundation models on
  low-level vision.
\newblock In \emph{ICLR}, 2024{\natexlab{a}}.

\bibitem[Wu et~al.(2024{\natexlab{b}})Wu, Zhang, Zhang, Chen, Liao, Wang, Xu,
  Li, Hou, Zhai, Xue, Sun, Yan, and Lin]{qinstruct}
Haoning Wu, Zicheng Zhang, Erli Zhang, Chaofeng Chen, Liang Liao, Annan Wang,
  Kaixin Xu, Chunyi Li, Jingwen Hou, Guangtao Zhai, Geng Xue, Wenxiu Sun, Qiong
  Yan, and Weisi Lin.
\newblock Q-instruct: Improving low-level visual abilities for multi-modality
  foundation models.
\newblock In \emph{CVPR}, pages 25490--25500, 2024{\natexlab{b}}.

\bibitem[Wu et~al.(2016)Wu, Li, Meng, Ngan, Luo, Huang, and Zeng]{ICLT}
Qingbo Wu, Hongliang Li, Fanman Meng, King~N. Ngan, Bing Luo, Chao Huang, and
  Bing Zeng.
\newblock Blind image quality assessment based on multichannel feature fusion
  and label transfer.
\newblock \emph{IEEE Transactions on Circuits and Systems for Video
  Technology}, 26\penalty0 (3):\penalty0 425--440, 2016.

\bibitem[Wu et~al.(2018)Wu, Li, Ngan, and Ma]{LOCRUE}
Qingbo Wu, Hongliang Li, King~N. Ngan, and Kede Ma.
\newblock Blind image quality assessment using local consistency aware
  retriever and uncertainty aware evaluator.
\newblock \emph{IEEE Transactions on Circuits and Systems for Video
  Technology}, 28\penalty0 (9):\penalty0 2078--2089, 2018.

\bibitem[Wu et~al.(2023)Wu, Sun, Zhu, Zhao, and Li]{HPS}
Xiaoshi Wu, Keqiang Sun, Feng Zhu, Rui Zhao, and Hongsheng Li.
\newblock Human preference score: Better aligning text-to-image models with
  human preference.
\newblock In \emph{IEEE/CVF International Conference on Computer Vision}, pages
  2096--2105, 2023.

\bibitem[Xu et~al.(2022)Xu, De~Mello, Liu, Byeon, Breuel, Kautz, and
  Wang]{GroupViT}
Jiarui Xu, Shalini De~Mello, Sifei Liu, Wonmin Byeon, Thomas Breuel, Jan Kautz,
  and Xiaolong Wang.
\newblock Groupvit: Semantic segmentation emerges from text supervision.
\newblock In \emph{CVPR}, pages 18113--18123, 2022.

\bibitem[Xu et~al.(2024{\natexlab{a}})Xu, Liu, Wu, Tong, Li, Ding, Tang, and
  Dong]{ImageReward}
Jiazheng Xu, Xiao Liu, Yuchen Wu, Yuxuan Tong, Qinkai Li, Ming Ding, Jie Tang,
  and Yuxiao Dong.
\newblock Imagereward: learning and evaluating human preferences for
  text-to-image generation.
\newblock In \emph{NIPS}, pages 15903--15935, 2024{\natexlab{a}}.

\bibitem[Xu et~al.(2024{\natexlab{b}})Xu, Liao, Xiao, Chen, Wu, Yan, and
  Lin]{LoDa}
Kangmin Xu, Liang Liao, Jing Xiao, Chaofeng Chen, Haoning Wu, Qiong Yan, and
  Weisi Lin.
\newblock Boosting image quality assessment through efficient transformer
  adaptation with local feature enhancement.
\newblock In \emph{CVPR}, pages 2662--2672, 2024{\natexlab{b}}.

\bibitem[Xu et~al.(2018)Xu, Zhang, Huang, Zhang, Gan, Huang, and He]{Attngan}
Tao Xu, Pengchuan Zhang, Qiuyuan Huang, Han Zhang, Zhe Gan, Xiaolei Huang, and
  Xiaodong He.
\newblock Attngan: Finegrained text to image generation with attentional
  generative adversarial networks.
\newblock In \emph{CVPR}, pages 1316--1324, 2018.

\bibitem[Yang et~al.(2024)Yang, Fu, Zhang, Cao, Liu, and Peng]{MoE-AGIQA}
Junfeng Yang, Jing Fu, Wei Zhang, Wenzhi Cao, Limei Liu, and Han Peng.
\newblock Moe-agiqa: Mixture-of-experts boosted visual perception-driven and
  semantic-aware quality assessment for ai-generated images.
\newblock In \emph{CVPRW}, pages 6395--6404, 2024.

\bibitem[Yang et~al.(2022)Yang, Wu, Shi, Lao, Gong, Cao, Wang, and
  Yang]{MANIQA}
Sidi Yang, Tianhe Wu, Shuwei Shi, Shanshan Lao, Yuan Gong, Mingdeng Cao, Jiahao
  Wang, and Yujiu Yang.
\newblock Maniqa: Multi-dimension attention network for no-reference image
  quality assessment.
\newblock In \emph{CVPRW}, pages 1190--1199, 2022.

\bibitem[Yao et~al.(2022)Yao, Huang, Hou, Lu, Niu, Xu, Liang, Li, Jiang, and
  Xu]{FILIP}
Lewei Yao, Runhui Huang, Lu Hou, Guansong Lu, Minzhe Niu, Hang Xu, Xiaodan
  Liang, Zhenguo Li, Xin Jiang, and Chunjing Xu.
\newblock Filip: Fine-grained interactive language-image pre-training.
\newblock In \emph{International Conference on Learning Representations}, 2022.

\bibitem[Yao et~al.(2023)Yao, Cao, Feng, Cheng, and Han]{oln}
Xiwen Yao, Qinglong Cao, Xiaoxu Feng, Gong Cheng, and Junwei Han.
\newblock Learning to assess image quality like an observer.
\newblock \emph{IEEE Transactions on Neural Networks and Learning Systems},
  34\penalty0 (11):\penalty0 8324--8336, 2023.

\bibitem[Ying et~al.(2020)Ying, Niu, Gupta, Mahajan, Ghadiyaram, and
  Bovik]{LIVEFB}
Zhenqiang Ying, Haoran Niu, Praful Gupta, Dhruv Mahajan, Deepti Ghadiyaram, and
  Alan~C. Bovik.
\newblock From patches to pictures (paq-2-piq): Mapping the perceptual space of
  picture quality.
\newblock In \emph{CVPR}, pages 3572--3582, 2020.

\bibitem[You and Korhonen(2021)]{triq}
Junyong You and Jari Korhonen.
\newblock Transformer for image quality assessment.
\newblock In \emph{ICIP}, pages 1389--1393, 2021.

\bibitem[Yu et~al.(2024)Yu, Guan, Lu, Li, and Chen]{SF-IQA}
Zihao Yu, Fengbin Guan, Yiting Lu, Xin Li, and Zhibo Chen.
\newblock Sf-iqa: Quality and similarity integration for ai generated image
  quality assessment.
\newblock In \emph{CVPRW}, pages 6692--6701, 2024.

\bibitem[Zhang et~al.(2017)Zhang, Xu, Li, Zhang, Wang, Huang, and
  Metaxas]{Stackgan}
Han Zhang, Tao Xu, Hongsheng Li, Shaoting Zhang, Xiaogang Wang, Xiaolei Huang,
  and Dimitris~N Metaxas.
\newblock Stackgan: Text to photorealistic image synthesis with stacked
  generative adversarial networks.
\newblock In \emph{ICCV}, pages 5907--5915, 2017.

\bibitem[Zhang et~al.(2024)Zhang, Huang, Jin, and Lu]{survey}
Jingyi Zhang, Jiaxing Huang, Sheng Jin, and Shijian Lu.
\newblock Vision-language models for vision tasks: A survey.
\newblock \emph{IEEE Transactions on Pattern Analysis and Machine
  Intelligence}, 46\penalty0 (8):\penalty0 5625--5644, 2024.

\bibitem[Zhang et~al.(2015)Zhang, Zhang, and Bovik]{ILNIQE}
Lin Zhang, Lei Zhang, and Alan~C. Bovik.
\newblock A feature-enriched completely blind image quality evaluator.
\newblock \emph{IEEE Transactions on Image Processing}, 24\penalty0
  (8):\penalty0 2579--2591, 2015.

\bibitem[Zhang et~al.(2023{\natexlab{a}})Zhang, Rao, and Agrawala]{Diffusion01}
Lvmin Zhang, Anyi Rao, and Maneesh Agrawala.
\newblock Adding conditional control to text-to-image diffusion models.
\newblock In \emph{CVPR}, pages 3836--3847, 2023{\natexlab{a}}.

\bibitem[Zhang et~al.(2020)Zhang, Ma, Yan, Deng, and Wang]{dbcnn}
Weixia Zhang, Kede Ma, Jia Yan, Dexiang Deng, and Zhou Wang.
\newblock Blind image quality assessment using a deep bilinear convolutional
  neural network.
\newblock \emph{IEEE Transactions on Circuits and Systems for Video
  Technology}, 30\penalty0 (1):\penalty0 36--47, 2020.

\bibitem[Zhang et~al.(2023{\natexlab{b}})Zhang, Li, Ma, Zhai, Yang, and
  Ma]{BIQA_CL}
Weixia Zhang, Dingquan Li, Chao Ma, Guangtao Zhai, Xiaokang Yang, and Kede Ma.
\newblock Continual learning for blind image quality assessment.
\newblock \emph{IEEE TPAMI}, 45\penalty0 (3):\penalty0 2864--2878,
  2023{\natexlab{b}}.

\bibitem[Zhang et~al.(2023{\natexlab{c}})Zhang, Zhai, Wei, Yang, and Ma]{LIQE}
Weixia Zhang, Guangtao Zhai, Ying Wei, Xiaokang Yang, and Kede Ma.
\newblock Blind image quality assessment via vision-language correspondence: A
  multitask learning perspective.
\newblock In \emph{CVPR}, pages 14071--14081, 2023{\natexlab{c}}.

\bibitem[Zhang et~al.(2023{\natexlab{d}})Zhang, Li, Sun, Liu, Min, and
  Zhai]{aigc1k}
Zicheng Zhang, Chunyi Li, Wei Sun, Xiaohong Liu, Xiongkuo Min, and Guangtao
  Zhai.
\newblock A perceptual quality assessment exploration for aigc images.
\newblock In \emph{Int. Conf. Multimedia and Expo. Worksh.}, pages 440--445,
  2023{\natexlab{d}}.

\bibitem[Zhong et~al.(2022)Zhong, Yang, Zhang, Li, Codella, Li, Zhou, Dai,
  Yuan, Li, and Gao]{RegionCLIP}
Yiwu Zhong, Jianwei Yang, Pengchuan Zhang, Chunyuan Li, Noel Codella,
  Liunian~Harold Li, Luowei Zhou, Xiyang Dai, Lu Yuan, Yin Li, and Jianfeng
  Gao.
\newblock Regionclip: Region-based language-image pretraining.
\newblock In \emph{CVPR}, pages 16772--16782, 2022.

\bibitem[Zhu et~al.(2021)Zhu, Hou, Chen, Xie, Lu, and Che]{TranSLA}
Mengmeng Zhu, Guanqun Hou, Xinjia Chen, Jiaxing Xie, Haixian Lu, and Jun Che.
\newblock Saliency-guided transformer network combined with local embedding for
  no-reference image quality assessment.
\newblock pages 1953--1962, 2021.

\end{thebibliography}
